\documentclass[lettersize, journal]{IEEEtran}

\usepackage{amsmath}
\usepackage{amssymb}
\usepackage{amsfonts}
\usepackage{cite}
\usepackage{subfigure}
\usepackage[pdftex]{graphicx}
\usepackage[acronym, nonumberlist]{glossaries}
\usepackage{url}
\usepackage{threeparttable}
\usepackage[final, commandnameprefix=ifneeded]{changes}

\newacronym{lstm}{LSTM}{Long-Short Term Memory}
\newacronym{gru}{GRU}{Gated Recurrent Unit}
\newacronym{rnn}{RNN}{Recurrent Neural Network}
\newacronym{sac}{SAC}{Soft Actor-Critic}
\newacronym{vae}{VAE}{Variational Autoencoder}
\newacronym{bev}{BEV}{Bird-Eye View}
\newacronym{cnn}{CNN}{Convolutional Neural Network}
\newacronym{rl}{RL}{Reinforcement Learning}
\newacronym{ddpg}{DDPG}{Deep Deterministic Policy Gradient}
\newacronym[longplural={Artificial Neural Networks}]{ann}{ANN}{Artificial Neural Network}
\newacronym{mdn}{MDN}{Mixture Density Network}
\newacronym{gan}{GAN}{Generative Adversarial Network}
\newacronym{mlp}{MLP}{Multi-Layer Perceptron}
\newacronym{imu}{IMU}{Inertial Measurement Unit}
\newacronym{dqn}{DQN}{Deep Q-Network}
\newacronym{mse}{MSE}{Mean Squared Error}
\newacronym{msssim}{MS-SSIM}{Multi-Scale Structural Similarity Index Measure}
\newacronym{a2c}{A2C}{Advantage Actor Critic}
\newacronym{ppo}{PPO}{Proximal Policy Optimization}
\newacronym{ae}{AE}{Autoencoder}
\newacronym{dae}{DAE}{Dynamic Autoencoder}
\newacronym{cdae}{CDAE}{Combined Dynamic Autoencoder}
\newacronym{CARNet}{CARNet}{\textbf{C}ombined dyn\textbf{A}mic autoencode\textbf{R} \textbf{Net}work}
\newacronym{dvae}{DVAE}{Dynamic Variational Autoencoder}
\newacronym{wm}{WM}{World Models}
\newacronym{mdnrnn}{MDN-RNN}{Mixture Density Recurrent Neural Network}
\newacronym{mpc}{MPC}{Model-Predictive Control}
\newacronym{il}{IL}{Imitation Learning}

\newcommand*\Bell{\ensuremath{\boldsymbol\ell}}
\newcommand*\circled[1]{\tikz[baseline=(char.base)]{\node[shape=circle,draw,inner sep=1pt] (char) {#1};}}

\begin{document}

\title{CARNet: A Dynamic Autoencoder for Learning Latent Dynamics in Autonomous Driving Tasks}

\author{Andrey Pak$^1$, \emph{Student Member, IEEE}, Hemanth Manjunatha$^1$, \emph{Student Member, IEEE}\\ 
Dimitar Filev$^2$, \emph{Fellow, IEEE}, 
Panagiotis Tsiotras$^1$, \emph{Fellow, IEEE} 
\thanks{Manuscript created May, 2022.}
\thanks{$^1$School of Aerospace Engineering, Georgia Institute of Technology, Atlanta, GA 30332-1050.
Atlanta, GA 30332-1050. $^2$Research \& Advanced Engineering, Ford Motor Company,
Dearborn, MI 48121. 
All correspondence should be addressed to Andrey Pak \mbox{(andrey.pak@gatech.edu).}}}

\maketitle

\begin{abstract}
	Autonomous driving has received a lot of attention in the automotive industry and is often seen as the future of transportation. 
	Passenger vehicles equipped with a wide array of sensors (e.g., cameras, front-facing radars, LiDARs, and IMUs) capable of continuous perception of the environment are becoming increasingly prevalent.
	These sensors provide a stream of high-dimensional, temporally correlated data that is essential for reliable autonomous driving. 
	An autonomous driving system should effectively use the information collected from the 
	various sensors in order to form an abstract description of the world and maintain situational awareness. 
	Deep learning models, such as autoencoders, can be used for that purpose, as they can learn compact latent representations from a stream of incoming data. 
	However, most autoencoder models process the data independently, without assuming any temporal interdependencies. 
	Thus, there is a need for deep learning models that explicitly consider the temporal dependence of the data 
	in their architecture. 
	This work proposes CARNet, a Combined dynAmic autoencodeR NETwork
	architecture that utilizes an autoencoder combined with a recurrent neural network to learn the current latent representation and, in addition, also predict future latent representations in the context of autonomous driving.
	We demonstrate the efficacy of the proposed model in both imitation and reinforcement learning settings using both simulated and real datasets.
	Our results show that the proposed model outperforms the baseline state-of-the-art model, 
	while having significantly fewer trainable parameters.
\end{abstract}

\glsreset{ae}

\section{Introduction}

\gls{rl} and \gls{il} have been gaining much traction in the area of autonomous driving as
a promising avenue to learn an end-to-end policy, which directly maps sensor observations to steering and throttle commands. 
However, approaches based on \gls{rl} or \gls{il} require many training samples collected from a multitude of sensors of different modalities, including camera images, LiDAR scans, and Inertial-Measurement-Units (IMUs). 
These data modalities generate high-dimensional data that are both spatially and temporally correlated. 
In order
to effectively use the information collected from the various sensors and develop an abstract description of the world, this high-dimensional data needs to be reduced to low-dimensional representations~\cite{lesort2018state}. 
To this end, \glspl{vae} have been extensively used in the past to infer low-dimensional \textit{latent variables}.
The \glspl{vae} are trained to operate on high-dimensional data through a probabilistic inference process.
However, in most autoencoder or \gls{vae} models, the data samples are processed independently from each other, 
without imposing any temporal dependencies, 
thus resulting in a loss of performance if the data are generated by a dynamic system.
Furthermore, modeling temporal dependencies in the data allows to construct a good \textit{prediction model}, which 
has been shown to benefit \gls{rl}/\gls{il} scenarios~\cite{werbos1987learning, silver2017predictron}. 
Empirical evidence also suggests that the human brain constructs an internal model of the world that is used to reason about possible futures (what/if/should happen) and then combines the outcome of this prediction with sensory input to form a belief model \cite{shadmehr2012biological}.
For application areas such as autonomous driving, the temporal characteristics of the data are more imperative, as most of the system components involved are dynamic. 
The problem is much more complex compared to processing static data one step at a time, as other agents (e.g., other vehicles and pedestrians) are also involved and interact with the learning agent.
Thus, there is a need for predictive latent representation models that explicitly consider the temporal dependencies of the data.

\begin{figure}[ht!]
    \centering
    \includegraphics[width=0.85\linewidth]{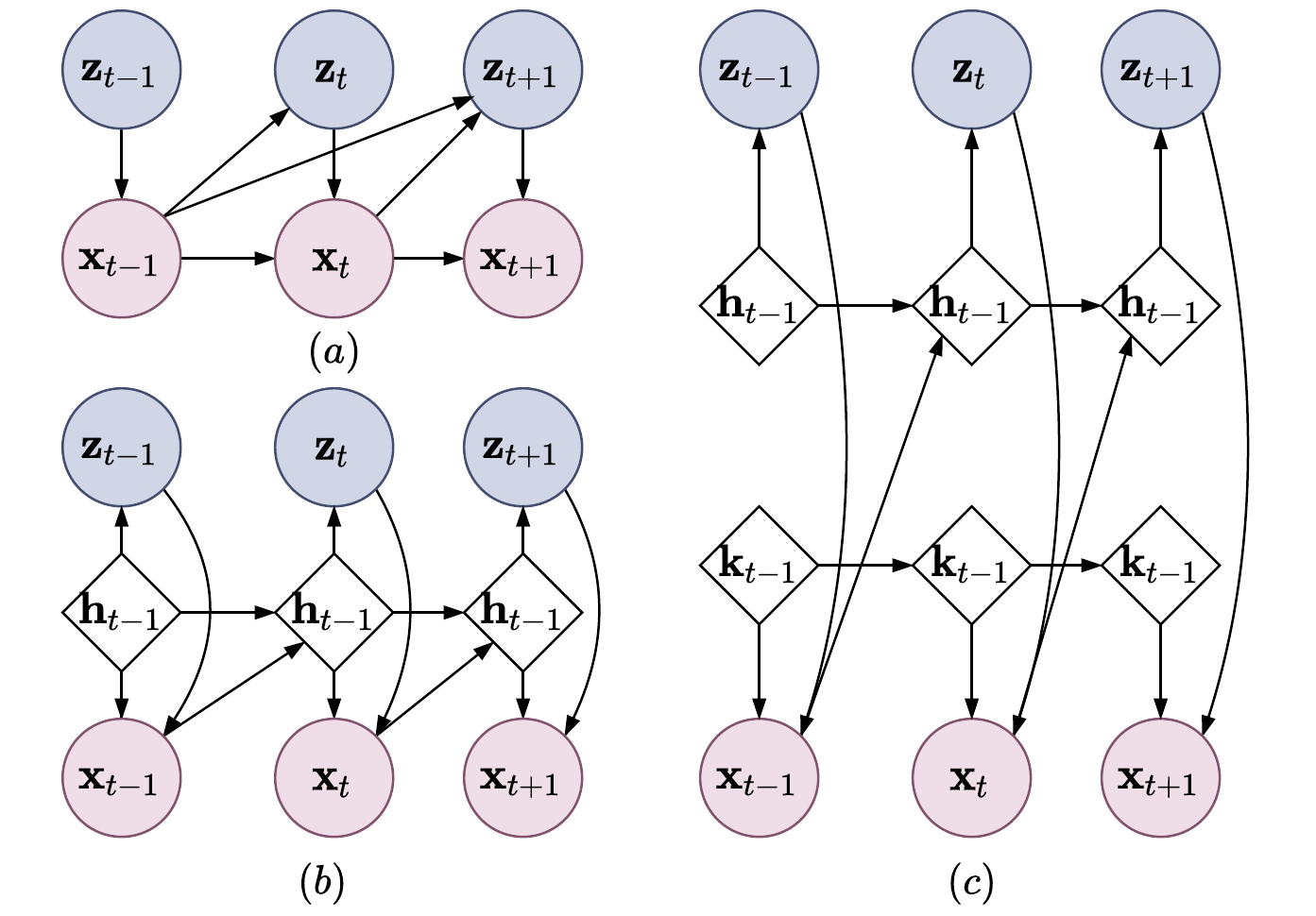}
    \caption{Two different implementations having same factorization (adapted from \cite{girin2020dynamical}). a) The probabilistic graphical model shows the dependency between observed vectors $\mathbf{x}_{1: T}$ and latent vectors $\mathbf{z}_{1: T}$ in terms of a recurrent neural network (RNN) implementation: b) Both $\mathbf{x}_{t}$ and $\mathbf{z}_{t}$ share a common hidden state $\mathbf{h}_{t}$, however, in c) the $\mathbf{x}_{t}$ and $\mathbf{z}_{t}$ are generated using two different hidden vectors, $\mathbf{k}_{t}$ and $\mathbf{h}_{t}$.}
    \label{fig:parameter-sharing}
\end{figure}

\glspl{dae} are suitable to model data that are inherently temporal \cite{girin2020dynamical}.
Given a sequence of random vectors $\mathbf{x}_{1: T}=\left\{\mathbf{x}_{t} \in \mathbb{R}^{n}\right\}_{t=1}^{T}$ and a sequence of latent random vectors $\mathbf{z}_{1: T}=\left\{\mathbf{z}_{t} \in \mathbb{R}^{n}\right\}_{t=1}^{T}$ 
that are assumed to be temporally correlated,
\glspl{dae} generate the joint distribution \textit{pdf} $p_{\theta}\left(\mathbf{x}_{1: T}, \mathbf{z}_{1: T}\right)$, where the $\theta$ parameters are usually estimated through deep learning models \cite{girin2020dynamical}. 
A particular \gls{dae} model can impose different conditional assumptions between  the observed vectors $\mathbf{x}_{1: T}$ and the latent vectors $\mathbf{z}_{1: T}$. 
For example, if we assume a first-order Markovian dependency, then $\mathbf{x}_{t}$ and $\mathbf{z}_{t}$ depend only on $\mathbf{x}_{t-1}$ and $\mathbf{z}_{t-1}$, respectively. 
For more general models,  one can accumulate the histories ($\mathbf{x}_{1:t-1}$ and $\mathbf{z}_{1:t-1}$) and use a Recurrent Neural Network (RNN) structure to learn the underlying cross-temporal dependencies~\cite{girin2020dynamical, ha2018recurrent}.

The imposed conditional assumptions do not always directly reflect the implementation details. 
For instance, consider the following conditional independence structure expressed using the two equations below
\begin{subequations} \label{eq:conditional-dependence}
\begin{align}
p\left(\mathbf{z}_{t} \mid \mathbf{x}_{1: t-1}, \mathbf{z}_{1: t-1}\right)&= p\left(\mathbf{z}_{t} \mid \mathbf{x}_{1: t-1}\right), \\
p\left(\mathbf{x}_{t} \mid \mathbf{x}_{1: t-1}, \mathbf{z}_{1: t}\right)&= p\left(\mathbf{x}_{t} \mid \mathbf{x}_{1: t-1}, \mathbf{z}_{t}\right).
\end{align}
\end{subequations}
Here it is assumed that $\mathbf{x}_{t}$ and $\mathbf{z}_{t}$ depend on the history $\mathbf{x}_{1: t-1}$ 
(this independence assumption is shown in Fig.~\ref{fig:parameter-sharing}a). 
Naturally, we can use an \gls{rnn} structure to capture this assumption, while completely differing in terms of the internal network implementation.
For example, in Fig.~\ref{fig:parameter-sharing}b) a single hidden state $\mathbf{h}_{t}$ is shared between $\mathbf{x}_{t}$ and $\mathbf{z}_{t}$. 
On the other hand, in Fig.~\ref{fig:parameter-sharing}c), two different hidden states, namely, 
$\mathbf{k}_{t}$ and $\mathbf{h}_{t}$ are used to model $\mathbf{x}_{t}$ and $\mathbf{z}_{t}$, respectively. 
Thus, the family of temporal models not only depends on the independence assumptions, but also on the implementation details \cite{girin2020dynamical}.

\begin{figure*}[t!] 
	\centering
    {\includegraphics[width=0.9\linewidth]{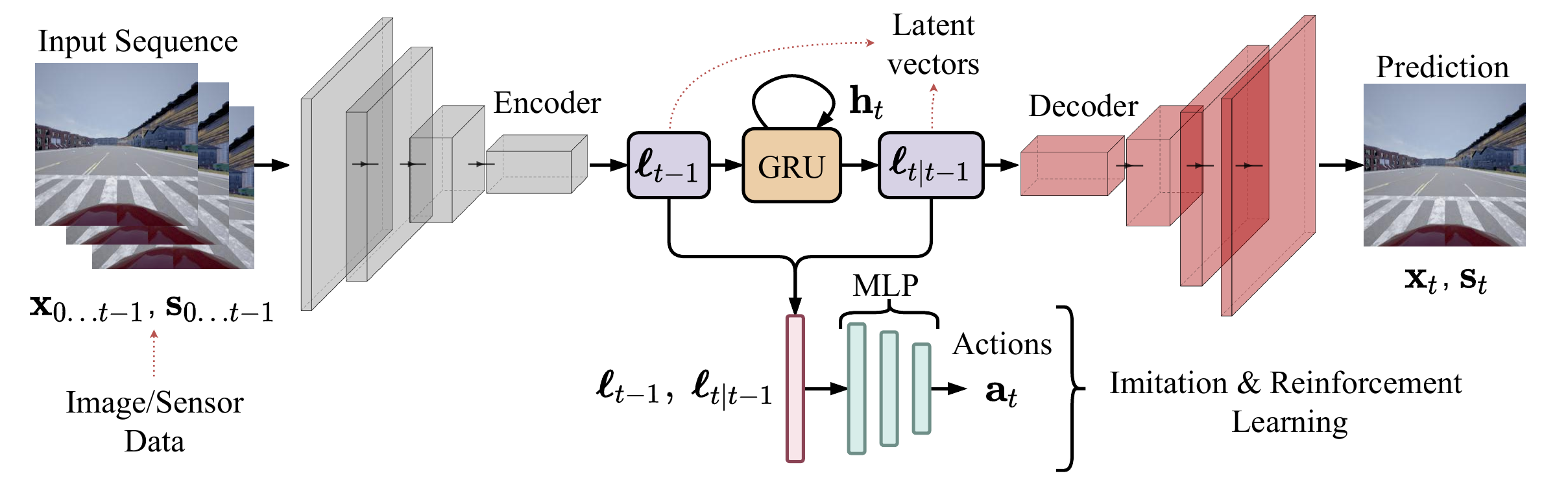}
	\caption{Overview of the \gls{CARNet} architecture with imitation and reinforcement learning. 
	The proposed \gls{CARNet} combines both the encoder and decoder networks with a recurrent neural network to learn present and future latent vectors. 
	These latent vectors are used to learn autonomous driving tasks using reinforcement learning.}
	\label{fig:overview}}
\end{figure*}

Motivated by this observation, we present a combined model for learning the latent space from high-dimensional, temporally correlated data.
The proposed combined architecture is shown in Fig.~\ref{fig:overview}. 
The architecture combines an \gls{ae} and an \gls{rnn} (in the form of \gls{gru}) in a single network with parameter sharing. 
Many previous works support the rationale for weight sharing \cite{girin2020dynamical, salimans2016structured, chung2015recurrent, kingma2019introduction}, where it has been shown that weight sharing between a generative model ($p(\mathbf{x}, \mathbf{z})$) and an inference model ($q(\mathbf{z}|\mathbf{x})$) can improve 
training. 
Throughout this paper, we refer to this combined model as \gls{CARNet}. 
The \gls{CARNet} learns to estimate the present latent vector and also predict the future latent vector from raw data. 
These latent vectors are then used to learn autonomous driving tasks using reinforcement learning.

In addition to the proposed combined architecture to learn latent space representations, we also explore the effect of attention mechanisms~\cite{vaswani2017attention} on latent vector estimation from images. 
The reasoning behind this idea is that not all parts of the input image are equally informative, and are not all equally relevant for planning. 
For instance, recent works \cite{tang2020neuroevolution, gelada2019deepmdp} have shown that an agent does not have to learn representations that are capable of reconstructing the complete observation; instead, it is sufficient that the representation allows predicting only the quantities that are directly relevant for planning. 
Ideally, a latent space representation should incorporate only the essential information for the current decision-making task while filtering out all the unnecessary details. 
For example, while driving, a driver's attention shifts toward an adjacent lane while performing a lane change. 
By focusing on the adjacent lane, the driver discards irrelevant objects and observations, thus reducing the unnecessary complexity of scene understanding, which facilitates efficient decision-making~\cite{chen2015deepdriving, ahlstrom2011processing}. 
We explore the use of attention to filter out unnecessary parts of the image for latent vector estimation. 

\subsection*{Contributions}

We present a new \textbf{C}ombined dyn\textbf{A}mic autoencode\textbf{R} \textbf{Net}work (CARNet) that learns current and future latent representations from raw high-dimensional data and, in addition, acts as a future latent vector prediction model.
In contrast to similar state-of-the-art end-to-end driving architectures (e.g., \cite{ha2018recurrent}), our approach has the benefits of sharing the parameters between the \gls{ae} and the \gls{rnn}, which is shown to increase 
performance. 
Due to the architecture of the proposed model, latent vector learning has additional continuity constraints, which enables a smooth transition between latent vectors. 
In addition, the proposed model has significantly fewer ($\approx$ 2.5 times) number of parameters than the selected baseline state-of-the art model. 
We validate the efficacy of the proposed model in two sets of experiments using imitation learning and reinforcement learning. 
The results show that the proposed model outperforms the baseline (e.g.,~\cite{ha2018recurrent}), while having significantly fewer parameters.
%
%

\section{Related Work}

Knowing how the environment may evolve in the immediate future is crucial for the successful and efficient planning of an agent interacting 
with a complex environment. 
Traditionally, this problem has been approached using \gls{mpc} along with system identification~\cite{swief2019survey}. 
While these classical approaches are powerful and well-established tools when the dynamical model is known and relatively simple, 
in practice it becomes increasingly difficult to apply these techniques to more realistic and complex problems, such autonomous driving.
Recently, driven by computer hardware advancements, along with the use of computational algorithms that harvest the expressivity and generalizability of deep neural network architectures,
it has become possible to learn representations of complex and realistic environments. 
This section gives an overview of the current approaches addressing learning, simulating, and predicting dynamic environments.

\glsreset{rnn} \glspl{rnn} are one of the most popular architectures frequently used to model time-series data. 
The idea of using \glspl{rnn} for learning the system dynamics and making future predictions, as well as for simulating previously observed environments, has recently gained popularity~\cite{wahlstrom2015pixels,watter2015embed,patraucean2016spatiotemporal,sun2016learning}.
The capability of encoding/generative models, combined with the predictive capabilities of the recurrent networks, 
provides a promising framework for learning complex environment dynamics. 
In this regard, Girin et al., \cite{girin2020dynamical} provide a broad overview of available methods that model the temporal dependencies within a sequence of data and the corresponding latent vectors. 
This overview covers the recent advancements in learning and predicting environments such as basic video games and frontal-camera autonomous driving scenarios based on the combination of an autoencoder and a recurrent network.


Generative recurrent neural network architectures have also been used in simulators.
In \cite{chiappa2017recurrent} the authors propose a recurrent neural network-based simulator that learns action-conditional based dynamics that can be used as a surrogate environment model while decreasing the computational burden. 
The RNN simulator was shown to handle different environments and capture long-term interactions. 
The authors also show that state-of-the-art results can be achieved in Atari games and a 3D Maze environment using a latent dynamics model that is learned from observing both human and AI playing.
The authors also highlighted the limitations of the proposed \gls{rnn} architecture for learning more complex environments (such as 3D car racing and 3D Maze navigation) and identified the existing trade-offs between long and short-term prediction accuracy. 
They also discussed and compared the implications of prediction/action/observation dependent and independent transitions. 

One of the most influential works along the lines of environment modeling for RL tasks is the World Models (WM) by
Ha and Schmidhuber~\cite{ha2018recurrent}.
They propose to model the world dynamics using \gls{vae} and \gls{rnn} in a reinforcement learning context. 
They showed how spatial and temporal features extracted from the environment in an unsupervised manner could be successfully used to infer the desired control input, and how they can yield a policy that is sufficient for solving typical \gls{rl} benchmarking tasks in a simulated environment. 
The proposed WM architecture consists of three parts. 
First, a \gls{vae} is used to learn the latent representation of several simulated environments (e.g.,
Car Racing and VizDoom).
Second, a \gls{mdn}-\gls{rnn} \cite{graves2014generating}  performs state predictions based on the learned latent variable and the corresponding action input. 
Finally, a simple \gls{mlp} determines the control action from the latent variables. 
The authors achieved state-of-the-art performance from training the model both in an actual environment and the simulated latent space, i.e. the ``dream'' generated by sampling the \gls{vae}'s probabilistic component. 
In VizDoom \cite{kempka2016vizdoom} the model, which was trained purely in the simulated environment (i.e., the ``dream"), was able to apply its experience to the real game environment and even take advantage of the inherent flaws of the game.
The central intuition behind the WM approach is that living beings develop an internal perception model of the outside world based on their sensory inputs, which is then used for everyday activities. 
The WM architecture inspired the proposed model in the sense that we also use a combination of an \gls{ae} and an \gls{rnn}. 
%
%
We improved the baseline WM approach by enabling parameter sharing, that has been shown \cite{girin2020dynamical, salimans2016structured, chung2015recurrent, kingma2019introduction} to increase the quality of learning the dynamic transitions. 
Additional losses are introduced to improve latent space continuity, and the overall size and complexity of the combined architecture is reduced.

In \cite{hafner2019learning} the same authors as in \cite{ha2018recurrent} addressed the challenge of learning the latent dynamics of an unknown environment. 
The proposed PlaNet architecture uses image observations to learn a latent dynamics model and then subsequently performs planning and action choice in the latent space. 
The latent dynamics model uses both deterministic and stochastic components in a recurrent architecture to improve the planning performance.

In a similar study from Nvidia, GameGAN \cite{kim2020learning} proposed a generative model that learns to visually imitate the desired game by observing screenplay and keyboard actions during training. 
Given the gameplay image and the corresponding keyboard input actions, the network learned to imitate that particular game. 
The approach in \cite{kim2020learning} can be divided into three main parts: 
the Dynamics Engine, an action-conditioned \gls{lstm} \cite{hochreiter1997long,chiappa2017recurrent} that learns action-based transitions; the external Memory Module, whose objective is to capture and enforce long-term consistency; and the Rendering Engine that renders the final image not only by decoding the output hidden state but also by decomposing the static and dynamic components of the scene, such as the background environment vs. moving agents. 
The authors achieve state-of-the-art results for some game environments such as Pacman and VizDoom~\cite{kempka2016vizdoom}.

The recent DriveGAN \cite{kim2021drivegan} approach proposes a scalable, fully-differentiable simulator that also learns the latent representation of a given environment. 
The latent representation part of the DriveGAN uses a combination of $\beta$-\gls{vae} \cite{higgins2016beta} for inferring the latent representation and StyleGAN \cite{karras2019style,karras2020analyzing} for generating an image from the latent vector. 
This approach has an advantage over classic \gls{vae}, as in this case a similarity metric between the produced images is also learned during the training process~\cite{larsen2016autoencoding}.
%
Like the WM, StyleGAN uses a pre-training step to learn the latent space. 
Afterwards, the network is used to learn the connection between the actual latent representations and the corresponding actions taken between the transitions using the custom Dynamic Engine architecture that learns the transitions between latent vectors.
The \gls{gan} component improves the image decoding qualities and can make sharper reconstructions compared to an exclusively convolutional decoder.
The last two \gls{gan}-based approaches mostly explore generational capabilities of the proposed architectures in simulating the future state based on previous image and action sequences. However, these architectures can be adapted directly in the context of \gls{rl} or imitation learning.

In summary, all current state-of-the-art techniques use two main components to perform accurate future predictions: 
first, an unsupervised pre-training step is utilized to learn the latent space of the given environment.
This step usually involves an autoencoder or an autoencoder-\gls{gan} combination; 
second, a recurrent network, usually in the form of \gls{lstm} is used to capture the spatio-temporal dependencies in the data and make future predictions based on the previously observed frames and action inputs. %
However, most of these prior works train the autoencoder and recurrent architectures separately, with one of the main reasons being the training speed. While this training approach is perfectly reasonable and achieves state-of-the-art results, the learned latent representation is static. Moreover, some dynamic relations captured by the recurrent part might not be rich-enough.
%
On the contrary, in this paper, we combine the autoencoder and recurrent architectures into a single architecture that is trained simultaneously and outperforms the baseline network where these parts are trained separately.

\label{sec:approach}

\section{Approach}
\begin{figure*}[t!] 
	\centering
    \includegraphics[width=\linewidth]{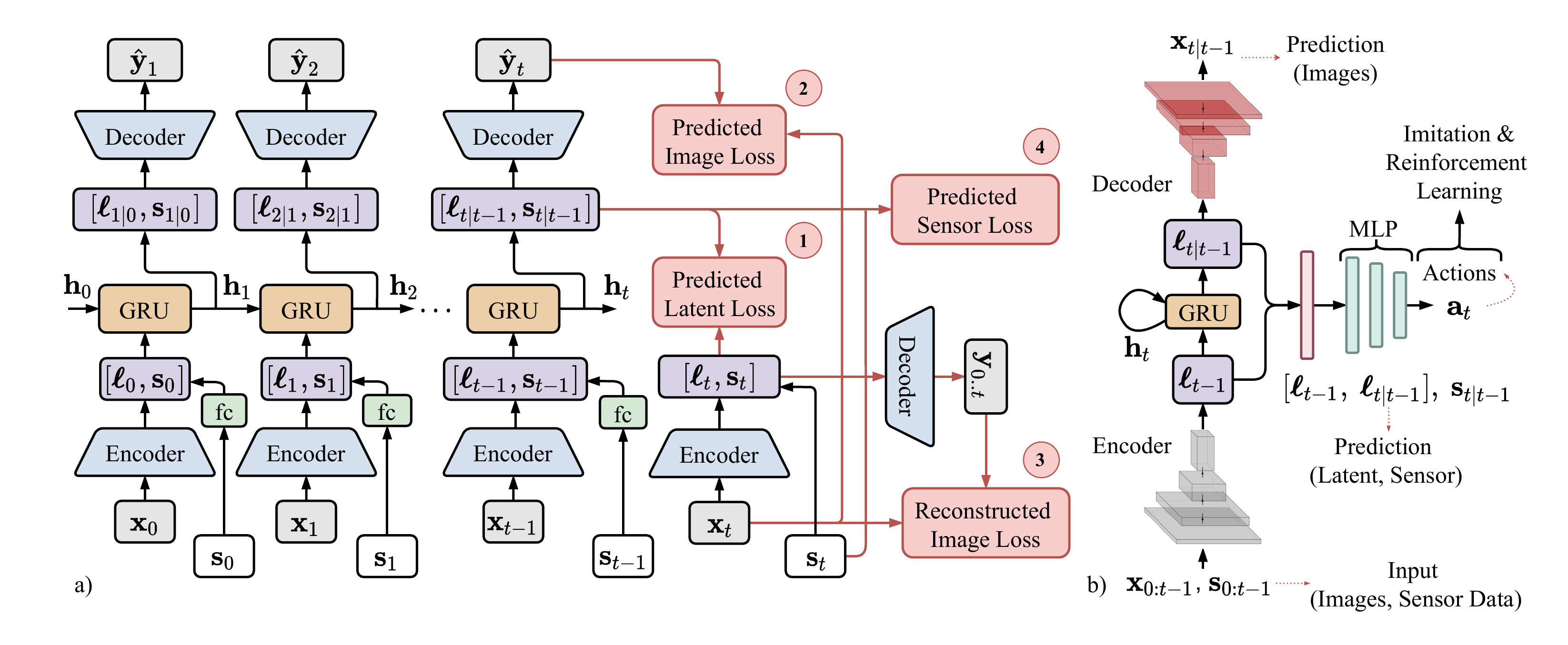}
	\caption{Combined recurrent architecture. Sequence of images is first encoded into latent space, fused (concatenated) with corresponding sensor measurements, then propagated through the recurrent network and outputs the estimate of both latent state and sensor measurements at timestep $t$.}
	\label{fig:combined_architecture_all}
\end{figure*}

Our main objective is to use the capability of a combined \gls{ae}-\gls{rnn} architecture to make accurate predictions of future states using latent representations.
The proposed approach is inspired by the World Models (WM) \cite{ha2018recurrent} architecture.
However, in contrast with WM, where the \gls{vae} and \gls{rnn} do not share any parameters, in our approach the 
\gls{ae} and \gls{rnn} networks are trained together. 
Also, instead of a \gls{vae}, we use an \gls{ae} architecture.
The first reason behind choosing an \gls{ae} rather than a \gls{vae} architecture
stems from our desire to
explore the minimum possible configuration of the network; adding a Gaussian mixture model to produce the latent variable would increase the number of parameters. 
The second reason is to avoid ``dream'' training that may lead to adversarial policies, where the probabilistic model ``cheats'' the environment by generating trajectories that do not follow the laws governing the actual environment~\cite{ha2018recurrent}.
In the context of autonomous driving, such trajectories are unacceptable for safety concerns.

\subsection{CARNet Architecture}
\label{sec:cdae_architecture}
The overall architecture of the proposed \gls{CARNet} is shown in 
Fig.~\ref{fig:combined_architecture_all}. 
The main intuition behind training together the \gls{ae} and \gls{gru} networks is as follows: 
while the autoencoder itself provides sufficient capabilities for capturing the latent state and reconstructing the subsequent images, 
it may omit temporal dependencies that are necessary for the proper prediction of the latent variable. 
Additionally, it is possible to further enforce temporal continuity in the latent space by introducing additional losses. 
In addition to operating purely on latent vectors, the architecture can also incorporate sensor measurements in the latent/hidden vectors (see Fig.~\ref{fig:combined_architecture_all}). 
%
The rationale is that images alone are not sufficient to capture complex environments (e.g., driving) 
and additional inputs may be needed.
Moreover, autonomous vehicles are already equipped with a wide array of sensors that can be used to sense the environment.

\begin{table}[h!]
    \renewcommand{\arraystretch}{1.25}
    \centering
    \caption{Autoencoder structure}
    \label{tab:autoencoder_architecture}
        \begin{tabular}{|c|c|c|c|}
        
        \multicolumn{2}{c}{\textbf{Encoder}}                                                                   & \multicolumn{2}{c}{\textbf{Decoder}}            \\ \hline
        \textbf{Layer}                                                       & \textbf{Output Shape} & \textbf{Layer}                                                         & \textbf{Output Shape}    \\ \hline
        \textbf{Input}                                                       & 1x256x256    & \textbf{Input}                                                         & 1x128     \\ \hline
        \begin{tabular}[c]{@{}l@{}}conv3-2\\ conv3-2\end{tabular}   & 2x128x128    & \begin{tabular}[c]{@{}l@{}}tconv3-64\\ tconv3-64\end{tabular} & 64x4x4    \\ \hline
        \begin{tabular}[c]{@{}l@{}}conv3-4\\ conv3-4\end{tabular}   & 4x64x64      & \begin{tabular}[c]{@{}l@{}}tconv3-32\\ tocnv3-32\end{tabular} & 32x8x8    \\ \hline
        \begin{tabular}[c]{@{}l@{}}conv3-8\\ conv3-8\end{tabular}   & 8x32x32      & \begin{tabular}[c]{@{}l@{}}tconv3-16\\ tconv3-16\end{tabular} & 16x16x16  \\ \hline
        \begin{tabular}[c]{@{}l@{}}conv3-16\\ conv3-16\end{tabular} & 16x16x16     & \begin{tabular}[c]{@{}l@{}}tconv3-8\\ tconv3-8\end{tabular}   & 8x32x32   \\ \hline
        \begin{tabular}[c]{@{}l@{}}conv3-32\\ conv3-32\end{tabular} & 32x8x8       & \begin{tabular}[c]{@{}l@{}}tconv3-4\\ tconv3-4\end{tabular}   & 4x64x64   \\ \hline
        \begin{tabular}[c]{@{}l@{}}conv3-64\\ conv3-64\end{tabular} & 64x4x4       & \begin{tabular}[c]{@{}l@{}}tconv3-2\\ tconv3-2\end{tabular}   & 2x128x128 \\ \hline
        conv3-128                                                   & 128x1x1      & tconv3-1                                                      & 1x256x256 \\ \hline
        \end{tabular}
\end{table}
The architecture of the baseline autoencoder (see Table~\ref{tab:autoencoder_architecture}) follows a simple/transposed convolution architecture with a kernel size of $3\times 3$. All convolution layers are followed with a batch normalization layer and a ReLU activation. With the dimension of the input image being $256\times 256$, the resulting size of the latent variable is $128$. 
During performance comparison, both \gls{CARNet} and \gls{wm} share identical layer structure, apart from the probabilistic part in \gls{vae}.

To better understand the flow of the processing induced by the proposed architecture (Fig.~\ref{fig:combined_architecture_all}), let us consider a sequence of $t+1$ consecutive frames $\left[\mathbf{x}_0, \ldots,\mathbf{x}_{t}\right]$ taken from the frontal camera of a moving vehicle, along with the corresponding sensor data. 
Note that the subscript $0$ refers to the beginning of the current sliding window, and not to the beginning of the dataset.
Given the first $t$ frames $\left[\mathbf{x}_0, \ldots,\mathbf{x}_{t-1}\right]$ and corresponding sensor data $\left[\mathbf{s}_0, \ldots, \mathbf{s}_{t-1}\right]$, our aim is to learn the latent representations $\left[\Bell_0, \ldots, \Bell_{t-1}\right]$ and predict $\mathbf{x}_{t}$, $\Bell_{t}$, and $\mathbf{s}_{t}$. 
Referring to Fig.~\ref{fig:combined_architecture_all}, and
for the sake of simplicity, let us consider a single-step prediction at time step $t-1$, without any sensor data. 
The image $\mathbf{x}_{t-1}$ is encoded into the latent variable as $\Bell_{t-1}=\mathbf{E}(\mathbf{x}_{t-1})$, where $\mathbf{E}$ denotes the encoder. 
It is important to mention that in the case where we only propagate the images (i.e., without sensor data), $\dim(\mathbf{h}_t) = \dim(\Bell_t)$. 
Also, note that the encoder/decoder can be any architecture or a typical convolutional neural network; hence, we assume a general architecture as follows
\begin{subequations} \label{eq:encoder-decoder}
\begin{align} 
    \Bell_{t-1} & = \mathbf{E}(\mathbf{x}_{t-1}), \\
    \mathbf{y}_{t-1} & = \mathbf{D}(\Bell_{t-1}), \\
    \hat{\mathbf{y}}_{t} & = \mathbf{D}(\Bell_{t|t-1}).
\end{align}
\end{subequations}
The encoded vector $\Bell_{t-1}$ is used as the input to the \gls{gru} block, whose output is the latent space vector at the next time step $\Bell_{t|t-1} = \mathbf{h}_t$, given by
    \begin{subequations}
\begin{align}
    z_{t} &= \sigma\left(W_z \left[\mathbf{h_{t-1}}, \, \Bell_{t-1}\right]\right), \label{eq:gru-a}\\
    r_{t} &= \sigma\left(W_r  \left[\mathbf{h_{t-1}}, \, \Bell_{t-1}\right]\right), \\
    \tilde{h}_t &= \tanh\left( W  \left[r_t \odot \mathbf{h_{t-1}}, \, \Bell_{t-1} \right]\right), \\
    \mathbf{h_{t}} &= (1-z_t) \odot \mathbf{h_{t-1}} + z_t \odot \tilde{h}_t \equiv \Bell_{t|t-1},
  \end{align}
\end{subequations}
where $W,\ W_z,\ W_r$ are  learnable weights, $\sigma$ is a sigmoid function, and $\odot$ is the Hadamard product. 
The predicted latent variable $\Bell_{t|t-1}$ is used as the hidden state for the next time step, as well as an input for reconstructing the image at the next time step, $\hat{\mathbf{y}}_{t} = \mathbf{D}(\Bell_{t|t-1})$, where $\mathbf{D}$ denotes the decoder 
(see Eq.~(\ref{eq:encoder-decoder})). 
It is important to note here that $\mathbf{h}_{t-1}$ also contains information about $\mathbf{h}_{t-2}$
Thus, the predicted latent vector $\Bell_{t}$ at time step $t$ not only depends on $\Bell_{t-1}$, but also on $\Bell_{t-2}$ 
as per Eq.~\eqref{eq:gru-a}, and in the case of image data only, no action or sensor conditioning is present.

In the case where sensor data is used, this data is fed through a single fully-connected layer and concatenated with the latent image encoding at each time step. 
The hidden vector is given as $\mathbf{h}~\triangleq~[\Bell, \, W_s\mathbf{s}]$ where $\Bell$ and $\mathbf{s}$ are latent and sensor data vectors and $W_s$ are weights  that scale the sensor data appropriately. 
To summarize, we train  the \gls{CARNet} network to model the following transitions:

\begin{enumerate}
    \item[a)] 
    In case of image-based latent representation only (Fig.~\ref{fig:combined_architecture_all}, omitting the 
    sensor input $\mathbf{s}$ and the loss \circled{4}), the model is trained to represent the transition $T(\Bell_{t}|\Bell_{t-1}, \mathbf{h}_{t-1})$, or, $T(\Bell_{t}|\Bell_{t-1}, \Bell_{t-1|t-2})$.
    
    \item[b)] 
    In case of an augmented latent representation (image + sensor, Fig.~\ref{fig:combined_architecture_all}), the model is trained to represent the transition $T(\Bell_{t}|\Bell_{t-1}, \mathbf{s}_{t-1}, \mathbf{h}_{t-1})$, or $T(\Bell_{t}|\Bell_{t-1}, \mathbf{s}_{t-1}, \left[\Bell_{t-2}, W_s\mathbf{s}_{t-2}\right])$ as $\mathbf{h}~\triangleq~[\Bell, \, W_s\mathbf{s}]$ where the sensor data is passed through a fully-connected layer designated as $W_s$ rather than being concatenated directly to the latent space. The main reasoning for not directly concatenating the latent vector ($\Bell$) with sensor data ($\mathbf{s}$) is that the sensor data might not be of same scale as the latent vector. Hence, we use a fully connected layer with weights $W_s$ to scale the sensor data appropriately.
\end{enumerate}

It is important to note that the modeled recurrent network transition is
different from the one in World Models (WM)~\cite{ha2018recurrent}. 
In WM, the \gls{mdnrnn} action-conditioned state transition is defined as $T(\mathbf{z}_{t}|\mathbf{a}_{t-1}, \mathbf{z}_{t-1}, \mathbf{h}_{t})$, where $\mathbf{z}_{t-1}$ and $\mathbf{z}_{t}$ are the previous and predicted latent states,
$\mathbf{a}_t$ is the set of previous actions, and $\mathbf{h}_{t-1}$ is the predicted hidden state of the \gls{mdnrnn}. 
On the other hand, 
we do not condition the latent state prediction on previous actions i.e., \gls{CARNet} learns $T(\mathbf{z}_{t}| \mathbf{z}_{t-1}, \mathbf{h}_{t})$ or $T(\mathbf{z}_{t}|\mathbf{s}_{t-1}, \mathbf{z}_{t-1}, \mathbf{h}_{t})$, in case sensor data is used alongside with camera images. 
However, action information can be added if necessary. 
In that case, the \gls{CARNet} model represents the transition $T(z_{t}|\mathbf{z}_{t-1}, \mathbf{h}_{t}, \mathbf{s}_{t-1}, \mathbf{a}_{t-1})$.

\subsection{Training}
\label{subs:training}
For training, the overall loss function for the single-step prediction can be broken down into three terms given as follows
%
    \begin{align}
    L_{\text{\tiny{total}}} & = \underbrace{
    \begin{aligned}
        \frac{1}{n+1}\sum_{t=0}^{n}\Big[L_{\text{\tiny{MS-SSIM}}}(\mathbf{x}_{t}, \mathbf{y}_{t})\Big] + \frac{1}{n}\sum_{t=1}^{n}\Big[L_{\text{\tiny{MS-SSIM}}}(\mathbf{x}_{t}, \hat{\mathbf{y}}_{t}) \Big]
    \end{aligned}
    }_{\text{Image reconstruction}} \nonumber \\ 
    & + \underbrace{\vphantom{\frac{1}{n}} \frac{1}{n}\sum_{t=1}^{n} L_{1_s} \left(\Bell_{t|t-1}, \Bell_{t}\right)}_{\text{Latent prediction}} 
    + \underbrace{\vphantom{\frac{1}{n}} \frac{1}{n}\sum_{t=1}^{n} L_{1_s} \left(\mathbf{s}_{t|t-1}, \mathbf{s}_{t}\right)}_{\text{Sensor prediction}},    \label{eq:loss}
    \end{align}
where $\Bell_{t}$, $\Bell_{t|t-1}$ denote the encoded latent variable at step $t$ and the conditioned latent prediction, $\mathbf{x}_t$, $\mathbf{y}_t$, $\hat{\mathbf{y}}_t$ denote the source and reconstructed image, respectively, 
where $\hat{\mathbf{y}}_t$ corresponds to the reconstructed image after the \gls{gru} (Fig. \ref{fig:combined_architecture_all}, loss \circled{3}) and $\mathbf{y}_t$ corresponds to ground truth (Fig. \ref{fig:combined_architecture_all}, loss \circled{2}). 
For sensor data, $\mathbf{s}_{t}, \mathbf{s}_{t|t-1}$ represent the ground truth and the predicted sensor qualities, respectively.

In terms of objective functions, a smooth $L_1$ ($L_{1_s}$) loss is used for the latent variable (Fig. \ref{fig:combined_architecture_all}, loss \circled{1}) and the sensor data (Fig. \ref{fig:combined_architecture_all}, loss \circled{4}). 
\gls{msssim}~\cite{wang2003multiscale} is used for image prediction to preserve reconstructed image structure and also serves as some form of regularization.
In our experiments, 
using \gls{mse} alone resulted in mode collapse and very blurry reconstructions. 
SSIM is a similarity measure that compares structure, contrast, and luminance between images. 
\gls{msssim} is a generalization of this measure over several scales~\cite{wang2003multiscale}, as follows
\begin{equation}
    L_{\text{\tiny{MS-SSIM}}} = \left[l_M(\mathbf{x}, \mathbf{y})\right]^{\alpha_M} \prod_{j=1}^{M} \left[c_j(\mathbf{x}, \mathbf{y})\right]^{\beta_j} \left[s_j(\mathbf{x}, \mathbf{y})\right]^{\gamma_j},
    \label{eq:ms-ssim}
\end{equation}
where $\mathbf{x}$ and $\mathbf{y}$ are the images being compared,  $c_j(\mathbf{x}, \mathbf{y})$ and $s_j(\mathbf{x}, \mathbf{y})$ are the contrast and structure comparisons at scale $j$, and the luminance comparison $l_M(\mathbf{x}, \mathbf{y})$ (shown in Eq.~\eqref{eq:ms-ssim-comparisons}) is computed at a single scale $M$, and
$\alpha_M$, and $\beta_j$, $\gamma_j$ ($j=1,\ldots,N$) are weight parameters that are used to adjust the relative importance of the aforementioned components, i.e., constrast, luminance, and structure. These parameters are left to the default implementation values as follows
\begin{subequations}    \label{eq:ms-ssim-comparisons}
\begin{align}
    l(\mathbf{x}, \mathbf{y}) &= \frac{2\mu_x \mu_y + C_1}{\mu_x^2 + \mu_y^2 + C_1}, \\
    c(\mathbf{x},  \mathbf{y}) &= \frac{2\sigma_x \sigma_y + C_2}{\sigma^2_x + \sigma^2_y + C_2}, \\
    s(\mathbf{x}, \mathbf{y}) &= \frac{\sigma_{xy} + C_3}{\sigma_x\sigma_y + C_3}.
\end{align}
\end{subequations}
In (\ref{eq:ms-ssim-comparisons}), $\mu$ and $\sigma$ are the mean and the variance of the image and are treated as an estimate 
of the luminance and contrast of the image. 
The constants $C_1,C_2,C_3$ are given by $C_1=(K_1L)^2$, $C_2=(K_2L)^2$, $C_3=C_2/2$, where $L$ is the dynamic range of the pixels and $K_{1}, K_2$, are two scalar constants. 
Additionally, not only the final predicted variable is used in the loss function, but all intermediate predictions of the recurrent network are also used in order to improve continuity in the latent space. 

\subsection{Separate and Ensemble Training} 

When pre-training the autoencoder separately on individual images, no temporal relations are captured in the latent space. 
This can potentially impact the quality of predictions produced by the \gls{rnn} network. 
Since both the autoencoder and the \gls{rnn} parts are trained together, we did not train the combined architecture from scratch but only pre-trained the autoencoder and then fine-tuned it as part of a bigger network. 
This resulted in significantly less time spent when training the combined architecture.

\subsection{Attention Mechanism}
\label{sec:attention-mechanism}

\begin{figure}[th!]
	\centering
	\includegraphics[width=\columnwidth]{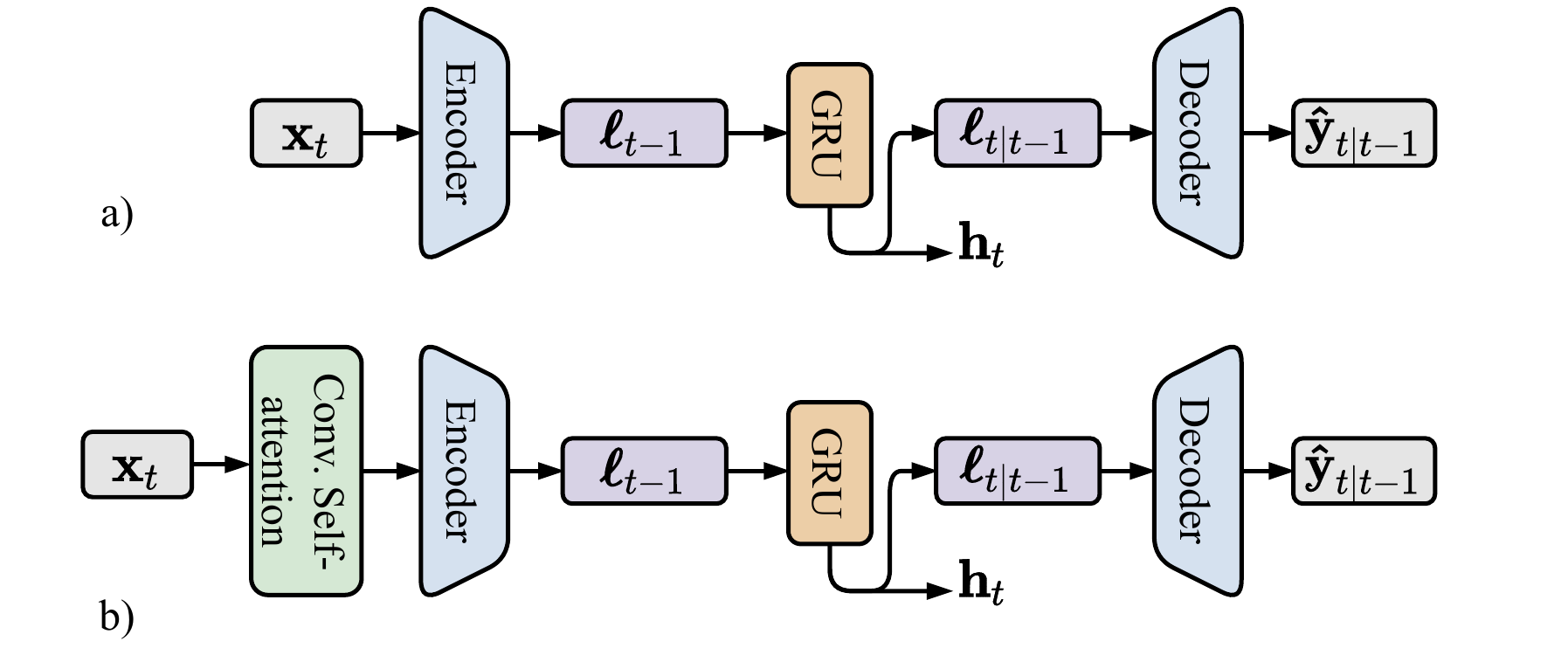}
	\caption{a) \gls{CARNet} without attention. b) \gls{CARNet} with convolution attention mechanism applied to input image.}
	\label{fig:attention_module}
\end{figure}

In addition to the \gls{ae}-\gls{rnn} architecture, we also implemented an attention mechanism \cite{vaswani2017attention} 
before the encoder (Fig. \ref{fig:attention_module}b). 
The motivation for the incorporation of the attention mechanism is that not all parts of the image carry 
relevant information for a particular task. 
Thus, by adding an attention module, we try to explore whether it results in a performance gain in the \gls{ae}-\gls{rnn} model. 
Since the input in the attention module is an image, we have used the self-attention mechanism for vision 
models~\cite{ramachandran2019stand}.

In terms of self-attention in the image space, given a pixel $x_{ij} \in \mathbb{R}^{d_{in}}$ of the input image $x$, a local region $a b \in \mathcal{N}_{k}(i, j)$ around the pixel is extracted with spatial extension $k$ centered at $x_{ij}$. 
The output pixel value $y_{ij}\in \mathbb{R}^{d_{out}}$ is calculated using a single-headed self-attention as follows

\begin{equation}       \label{eq:cnn-attention}
    y_{i j}=\sum_{a, b \in \mathcal{N}_{k}(i, j)} \operatorname{softmax}_{a b}\left(q_{i j}^{\top} k_{a b}\right) v_{a b},
\end{equation}
where, $q_{i j}=W_{Q} x_{i j}$, $k_{a b}=W_{K} x_{a b}$, and the values $v_{a b}=W_{V} x_{a b}$ are query, key, and values respectively, with learnable weights $W_{Q}, W_{K}, W_{V}$. 
The present formulation can be extended to multi-head attention, with each head having its own learnable parameters by partitioning the pixel features $x_{ij}$ depthwise into $N$ groups. 
Moreover, the formulation in Eq.~(\ref{eq:cnn-attention}) does not consider position information to encode attention, which limits the expressiveness in vision tasks. 
To remedy this, relative attention can be implemented that takes the relative position between the pixel $x_{ij}$ and each element in the region $a b \in \mathcal{N}_{k}(i, j)$. 
Thus, the new formulation is given as
\begin{equation}       \label{eq:cnn-relative-attention}
    y_{i j}=\sum_{a, b \in \mathcal{N}_{k}(i, j)} \operatorname{softmax}_{a b}\left(q_{i j}^{\top} k_{a b}+q_{i j}^{\top} r_{a-i, b-j}\right) v_{a b},
\end{equation}
where $r_{a-i, b-j}$ is the concatenated vector (i.e., $r_{a-i, b-j} = [r_{a-i},\ r_{b-i}]$) with row and column position information. Note that the dimension of $r_{a-i}$ and $r_{b-i}$ is $\frac{1}{2}d_{\text{out}}$, thus the dimension of $r_{a-i, b-j}$ is $d_{\text{out}}$.
More details on self-attention in vision tasks can be found in \cite{ramachandran2019stand}. 


\subsection{Simulated Data}

In order to learn good latent state representations from image data, a significant amount of data is required. 
Therefore, the frontal, side, and rear camera images are also included for regularizing and increasing data diversity (e.g., the side camera image becomes similar to the frontal camera image once the vehicle makes a right-angle turn). 
It is important to note that only the autoencoder part was pre-trained on all camera feeds, and the ensemble network only used the frontal camera. 
This proved helpful when using real-world datasets with misaligned data, i.e., with the absence of global shutter and mismatching data timestamps.
In terms of sensor measurements $\mathbf{s}_t$, we used steering, brake, and throttle data. 

\begin{figure}[th!]
	\centering
	\includegraphics[width=0.5\columnwidth]{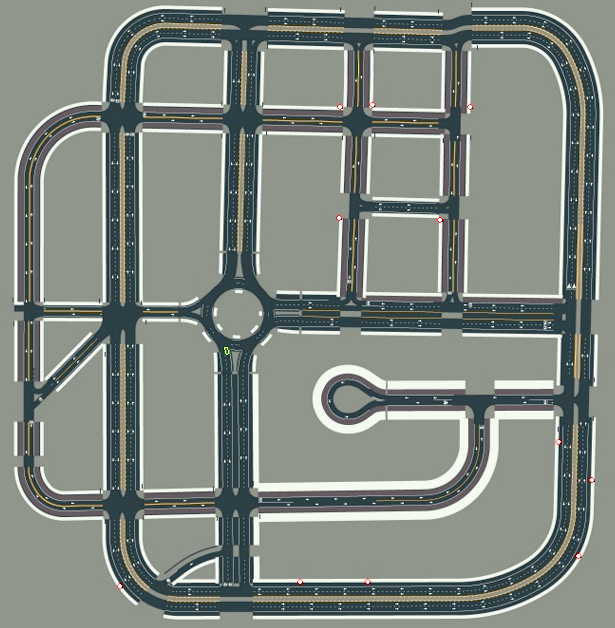}
	\caption{Map of the town used in CALRA for data generation.}
	\label{fig:town}
\end{figure}

The training data was generated using the CARLA \cite{dosovitskiy2017carla} simulator. 
A total of 300K time steps were generated using random roll-outs (random starting and goal points on the map shown in Fig.~\ref{fig:town}) utilizing the internal CARLA vehicle autopilot. 
The simulated data includes 4-directional camera images (front/left/right/rear) along with their corresponding semantic segmentation, IMU and other sensor data (speed, steering, LIDAR, etc.), desired control values, and additional experimental data for auxiliary tasks such as traffic light information.

\section{Experiments}

We considered two different cases to demonstrate the efficacy of the proposed architecture over the state-of-the-art architectures (e.g.,~\cite{ha2018recurrent}).
First, we considered an imitation learning case in which the task is to predict the autopilot actions given the history of states (i.e., a stack of images from the front view camera) for both simulated and real datasets. 
Imitation learning is cast as a nine-class classification problem where each class corresponds to a unique steering angle and acceleration combination. 
Note that for imitation learning, neither  \gls{CARNet} nor \gls{wm} are conditioned on the action history (see Section~\ref{subs:training}), as it may lead to bias in the classification. 
%
In the second case, we tested the proposed architecture in the CARLA simulator, which provides realistic scenarios in terms of visualization and traffic. 
For the second \gls{rl} scenario, \gls{CARNet} and \gls{wm} are conditioned on the sensor and action history accordingly (see explanation at the end of Section~\ref{sec:cdae_architecture}).

\subsection{Imitation Learning} \label{sec:imitation_learning}

%
Annotated data from expert demonstrations have been core drivers in the development and evaluation of learning systems, ranging from computer vision and natural language processing to robotic manipulation tasks and audio processing \cite{griffith2013policy, christiano2017deep}. 
The influence of such annotated data is significant in the field of \textit{Imitation Learning}, particularly in game-based platforms. 
The expert (usually human) demonstrations in such platforms have helped avoid the huge cost of learning from scratch \cite{silver2016mastering, vinyals2019grandmaster} and sometimes even achieve human-level  performance. 
Along the same line of work, we used expert demonstrations to train a neural network in a supervised manner and used the trained neural network as a starting point (``warm starting'') in a reinforcement learning experiment described in Section~\ref{sec:RLexp} below.

For imitation learning, given $N$ autopilot driving sequences $AP_i, i \in (1,\dots, N)$ with corresponding observation frames $I_{i,t}$, we learn a deterministic policy using a network parameterized by $\theta$ to mimic the autopilot actions.
The action space $A_{i,t}$ contains two continuous actions: steering angle and acceleration. 
However, for imitation learning, we discretized the continuous values into three levels. 
The steering angle assumes the values $[-0.2, 0.0, 0.2]\ \text{rad}$ and the acceleration 
command assumes the values $[-3, 0.0, 3]\ \text{m/s}$. 
Imitation learning is set up as a classification problem with nine classes resulting from different combinations of the previous discrete steering and acceleration values.
\added[id=HM]{Discretization of the action space is justifiable from a practical perspective, 
since in autonomous driving applications the high level decision policy derived by imitation learning, reinforcement learning, game theory or some other decision-making techniques is combined with low level trajectory planning and motion control algorithms \cite{nageshrao2019autonomous}.}
For training, we used mean cross-entropy loss between the predicted class and the autopilot class, as follows 
\begin{equation}  \label{eq:cross_entropy}
-\sum_{n=1}^N\sum_{k=1}^{K} w_{k} \log \frac{e^{ \left(\mathbf{x}_{n, k}\right)}}{\sum_{i=1}^{K}e^{\left(\mathbf{x}_{n, i}\right)}} \mathbf{y}_{n, k},
\end{equation}
where $N$ is the total number of autopilot driving sequences, and $K$ is the number of classes. 

\begin{figure}[th!]
	\centering
	\includegraphics[width=\columnwidth]{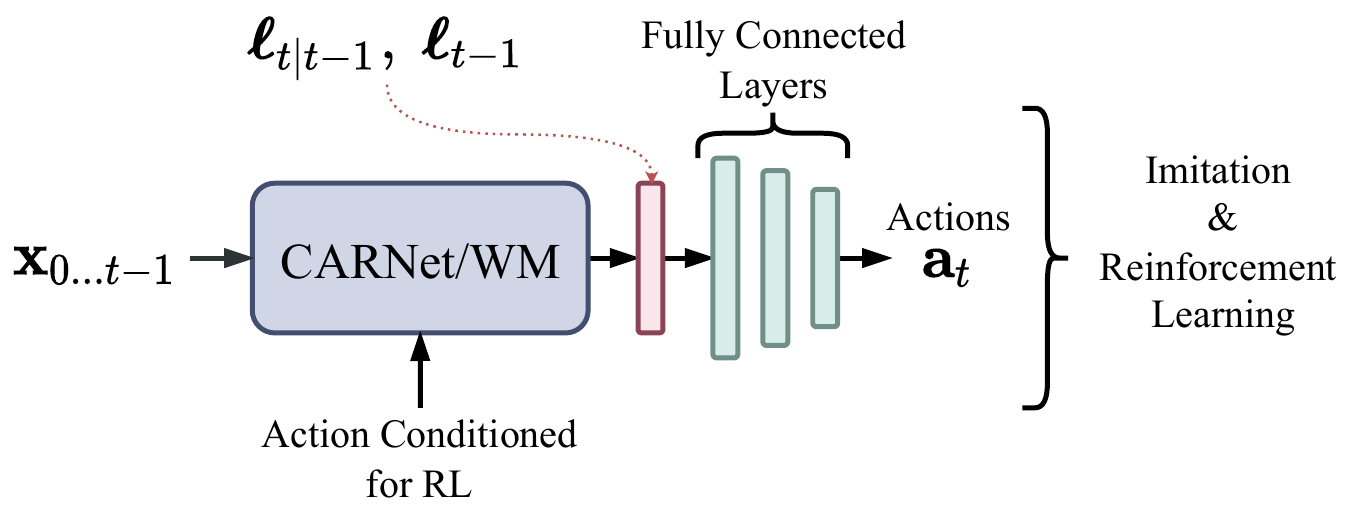}
	\caption{For imitation learning, \gls{CARNet} and \gls{wm} are not conditioned on history of actions. However, for reinforcement learning, both the models are conditioned on  the history of actions.}
	\label{fig:experimental-setup}
\end{figure}

The controller architecture for the imitation learning task
consists of a simple four-layer \gls{mlp} that gradually reduces the dimensionality of the latent space onto the action space (see Fig.~\ref{fig:experimental-setup}). 
The structure of the controller is given in Table~\ref{tab:controller_architecture}. 
The controller network uses the stacked previous and predicted latent vectors, so the input tensor size 
is twice the chosen latent size $\mathbf{a}_{} = \mathbf{C}([\Bell_{t-1}, \, \Bell_{t}])$, e.g., for a latent size of $128$, the size of the input tensor to the imitation learning network is $256$.

\begin{table}[ht!]
\renewcommand{\arraystretch}{1.25}
\centering
\caption{Controller structure (\gls{CARNet})}
\label{tab:controller_architecture}
\begin{tabular}{ccc}
\hline
\textbf{Layer} & \textbf{Input Size} & \textbf{Output Size} \\ 
\hline
FC1 + ReLU & 256 & 128    \\ 
FC2 + ReLU & 128 & 128    \\ 
FC3 + ReLU & 128 & 64    \\ 
FC4 + ReLU & 64 & 9    \\ \hline
\end{tabular}
\end{table}

The training was performed on a computer equipped with a GeForce RTX3090 GPU, Ryzen 5950x CPU, and 16GB of RAM. 
Also, the 300K time-steps are split into 70\% training, 15\% validation, and 15\% testing data.
The code implementation\footnote{\texttt{\url{https://github.com/apak-00/cdae}}} uses Pytorch with Adam optimizer.

\subsection{Baseline Comparison}

\begin{figure}[th!]
	\centering
	\includegraphics[width=0.75\columnwidth]{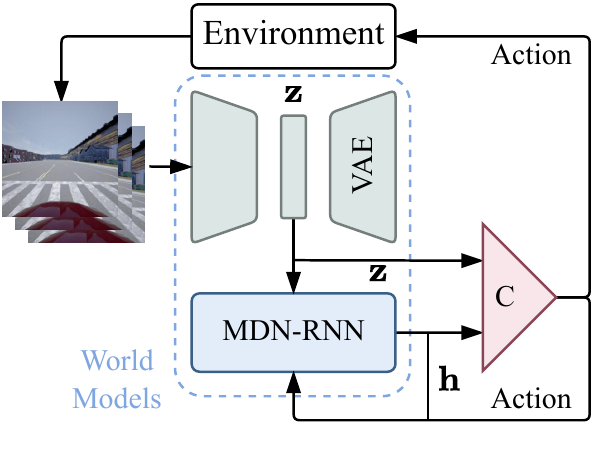}
	\caption{Flow diagram of the World Models architecture.}
	\label{fig:world-models}
\end{figure}

Since we drew our inspiration from dynamic autoencoders in general, and World Models in particular, 
\gls{wm} is chosen as a baseline comparison network. 
In \gls{wm}, the architecture consists of three different parts: a \gls{vae}, an MDN-RNN, and a fully connected controller network, which are all trained separately in a given order with the previous part of the network frozen. 
An overview of the WM network is shown in Fig.~\ref{fig:world-models}.

In \gls{wm}, the purpose of the mixture density network is to propagate the Gaussian Mixture that represents the learned latent state of the autoencoder and predict the hidden state. 
The \gls{mdnrnn} learns the transition
$P(\mathbf{h}_{t} | \mathbf{a}_{t-1}, \mathbf{z}_{t-1}, \mathbf{h}_{t-1})$ that predicts the next hidden state $\mathbf{h}_{t+1}$ that is concatenated with the previous latent encoding $\mathbf{z}_t$ for controller input. 
The output action produced is defined as $\mathbf{a}_{t} = W_c[\mathbf{z}_{t-1}, \, \mathbf{h}_{t}] + \mathbf{b}_c$.

We adapt the \gls{wm} architecture for the new scenario. First, the autoencoder part is modified to accommodate different image sizes, and more convolutional layers (apart from the reparametrization part) since the original input for the \gls{wm} included $64\times 64$ game screenplay, while we used $256\times 256$ camera images.
Both \gls{wm} and \gls{CARNet} autoencoders share identical convolutional encoder/decoder structure for fair comparison.
Second, since this paper is focused on exploring latent representations, the action-conditioning is removed from the recurrent predictions for the case of imitation learning, so that the recurrent part models the transition $T(\mathbf{z}_{t+1} | \mathbf{z}_t, \mathbf{h}_{t+1})$. 
Introducing action conditioning would bias the model toward learning the action dynamics, and it would be hard to show the improvements in the quality of the learned latent dynamics (see Section~\ref{subs:training}).
The controller parts of both networks are mostly identical layer-wise, with the only difference being the size of the input layers since \gls{wm} uses hidden vectors of larger size --- the hidden size of \gls{mdnrnn} that propagates the Gaussian mixture is $256$, so the size of the input tensor to the controller network is $128+256=384$.

\subsection{Reinforcement Learning}  \label{sec:RLexp}

We have carried out the data collection and reinforcement learning training in the CARLA environment~\cite{dosovitskiy2017carla}. 
We modified the OpenAI Gym wrapper \cite{chen2020interpretable} for the CARLA simulator and used the customized version as an interface for the environment required for generating data for our learning algorithms.
The reward function used for training is given as
\begin{equation}
    \begin{aligned}
        r = 200 \, r_{\rm collision} & + v_{\rm lon} + 10\frac{S}{S_\text{des}} \, r_{\rm fast} \\
        & + 40 \, r_{\rm out} -5 \alpha^2 + 0.2 \, r_{\rm lat} -0.1,
    \end{aligned}
\end{equation}
where $r_{\rm collision}$ is set to $-1$ if the vehicle collides; else, it is set to $0$; $v_{\rm lon}$ is the magnitude of the
projection of the vehicle velocity vector along the direction connecting the vehicle and the nearest next way-point;
$r_{\rm fast}$ is $-1$ when the car is faster than the desired speed else it is 0; $r_{\rm out}$ is $-1$ when the vehicle is out of the lane; $\alpha$ is the steering angle; $r_{\rm lat}$ refers to the lateral acceleration = $\alpha v^2$; the constant $-0.1$ is to make sure that the vehicle does not remain standstill. 
The penalty $r_{\rm fast}$ is either 0 or $-1$, which acts as a constraint for over-speeding.
Note that the coefficient for $r_{\rm out}$ is relatively large, as going out-of-lane causes a termination of the episode. 
Thus, the negative reward is only experienced once during an episode. 
Furthermore, a weight factor derived from normalizing the vehicle's speed $S$ by the desired speed $S_{\rm des} (30~\rm Km/h)$ is added to $r_{\rm fast}$; 
otherwise, the reward function would encourage a full-throttle policy. 
The action space is the same as in the imitation learning case (see Section~\ref{sec:imitation_learning}) i.e., nine discrete actions corresponding to different combinations of steering ($[-0.2, 0.0, 0.2]\ \text{rad}$) and acceleration ($[-3, 0.0, 3]\ \text{m/s}$).
Since the action space is discrete, we have used the DQN~\cite{mnih2013playing} algorithm for training the RL model. 
The hyper-parameters for the DQN model are shown in Table~\ref{tab:dqn_hyper_parameters}.

\subsection{Real World Data Evaluation}

For the real-world evaluation, we have used the Udacity dataset~\cite{udacity2018}. 
In order to match the simulated dataset scenario and structure, the extracted images were resized to $256\times 256$ dimensions and converted to grayscale. 
Since the Udacity dataset does not have a global shutter, resulting in camera and sensor timestamp mismatch, front camera timestamps were chosen as the reference ones, and closest images and sensor data were selected in relation to that timestamp. We post-processed throttle, steering, and brake values that were extracted from the rosbag file.
The post-processing step is as follows: brake and throttle were discretized to match the CARLA scenario,  in the range $[0, 1]$, and the steering wheel angle was normalized by dividing it by the steering wheel ratio of the car used in dataset collection, resulting in a normalized range of $[-1, 1]$.

\section{Results}

This section presents the results from two different experiments, corresponding to imitation and reinforcement learning.
In the first case, we explore autonomous driving in an imitation learning setting, which involves predicting the vehicle's actions (acceleration, braking, and throttle) from a sequence of four images from a front-facing camera. 
In the second scenario, we explore realistic autonomous driving scenarios in the CARLA environment \cite{dosovitskiy2017carla} within the reinforcement learning setting.
In both scenarios, the learned latent vectors are used to predict the vehicle's actions. 
Finally, we compare the performance of \gls{CARNet} and \gls{wm} in real-world data using Udacity dataset~\cite{udacity2018}.

\subsection{Ablation Analysis}

We performed ablation studies to select the best network configuration and hyper-parameters. Ablation studies were performed on the simulated dataset. The parameters of the ablation study are summarized in Table \ref{tab:ablation_parameters} and selected parameters are highlighted in bold. 

\begin{table}[!ht]
\renewcommand{\arraystretch}{1.5}
  \centering
  \caption{Ablation Studies Overview}
    \begin{tabular}{cccc}
    \hline
    \textbf{Parameters} & \textbf{Latent Size} & \textbf{RNN Unit} & \textbf{Loss Function}\\
    \hline
    \textbf{Range} & [64, \textbf{128}, 256] & [LSTM, \textbf{GRU}] & [MSE, \textbf{MS-SSIM}]\\
    \hline
    \end{tabular}
  \label{tab:ablation_parameters}
\end{table}%

\begin{figure}[!ht]
	\centering
	\includegraphics[width=\columnwidth]{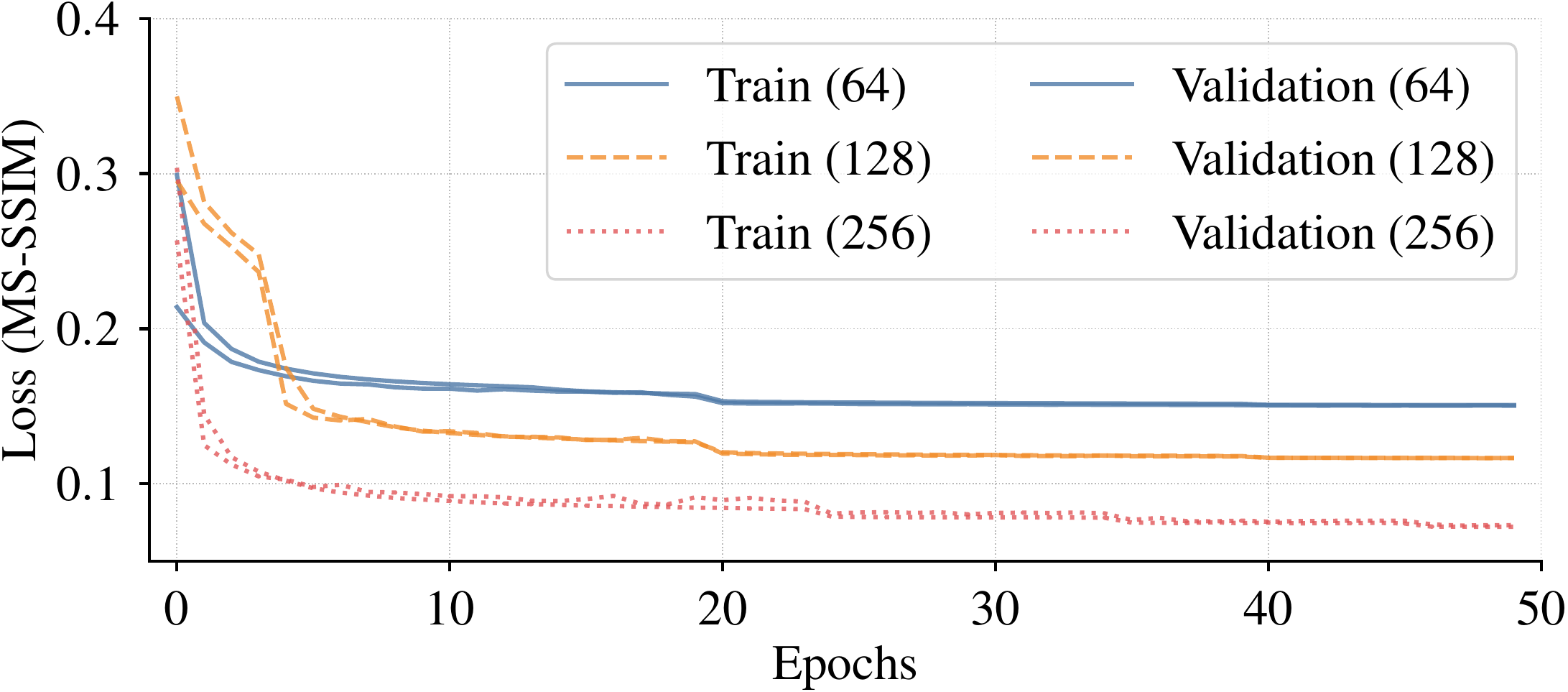}
	\caption{MS-SSIM loss for autoencoder reconstructions of various latent variable sizes. Note that the range for the \gls{msssim} loss is $[0, 1]$, where 0 corresponds to perfect structure match, while 1 shows complete mismatch, accordingly. A nearly linear relation can be observed between the latent variable size and final reconstruction quality. }
	\label{fig:ablation_ae}
\end{figure}

\begin{figure}[!ht]
	\centering
	\includegraphics[width=0.95\columnwidth]{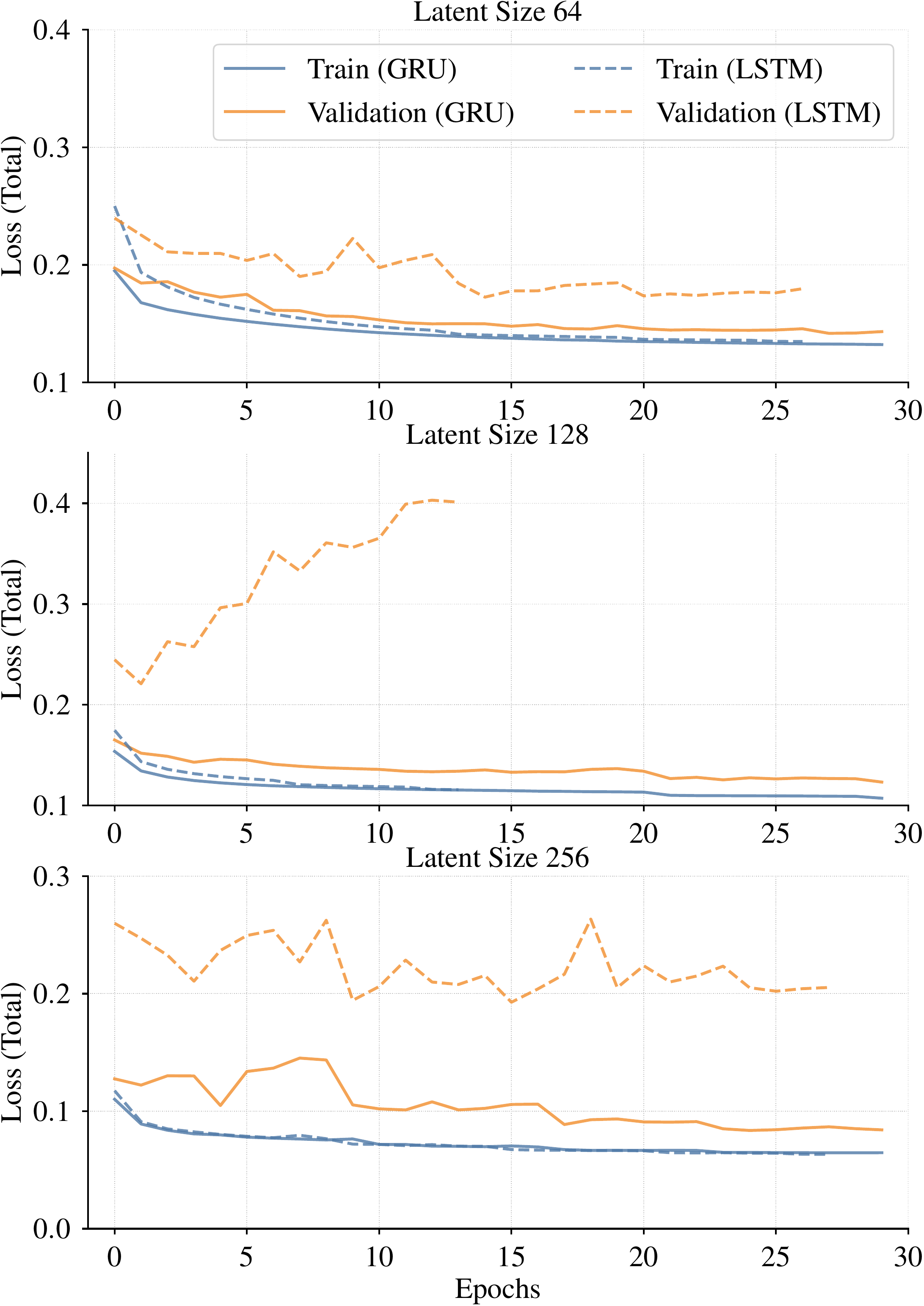}
	\caption{\gls{lstm} vs. GRU varying latent variable size training and validation curves. Note that \gls{lstm} training often triggered early stopping criteria (no decrease in validation loss, even with learning rate scheduling and showed high tendency to overfitting compared to \gls{gru}. Both \gls{rnn} architectures were trained under the same set of hyperparameters.}
	\label{fig:ablation_lstm_gru}
\end{figure}

First, during the initial experiments \gls{lstm} was replaced in favor of \gls{gru}, as the former showed heavy over-fitting behavior (see Fig. \ref{fig:ablation_lstm_gru}). 
Second, we selected \gls{msssim} for the image reconstruction loss as \gls{mse} often resulted in mode collapse and blurry image reconstruction since it did not impose any structural constraints on the decoded image. 
Lastly, we tested various latent sizes for the autoencoder and hidden vector sizes for \gls{rnn}. 
While reducing the latent size did address the over-fitting problem, reducing the size further led to low representation capacity (summarized in Fig. \ref{fig:ablation_ae}). 
Hence, we chose a latent size of 128. 
Moreover, in imitation learning experiments, the performance difference between 128 and 256 latent sizes appeared to be negligible (see Section~\ref{sec:imitation_learning}). 
Thus, a latent size of 128 provides a good trade-off between the representation capacity and the number of trainable parameters.

\begin{table}
    \begin{minipage}{.45\linewidth}
        \centering
        \setlength\tabcolsep{1.25pt}
        \renewcommand{\arraystretch}{1.10}
        \caption{Learning Hyperparameters}
        \vspace{-8pt}
        \begin{tabular}[t]{lc}
            \hline
            \textbf{Hyperparameter} & \textbf{Value}\\
            \hline
            Input image size & 256x256 \\
            Autoencoder latent size & 128 \\
            \gls{gru} hidden size & 128 \\
            RNN time-steps & 4\\
            Optimizer & Adam\\
            Learning rate (scheduled) & $10^{-3}$\\
            Batch size & 64\\
            \hline
        \end{tabular}
        \label{tab:learning_hyper_parameters}%
    \end{minipage}%
    \hfill%
    \begin{minipage}{.45\linewidth}
        \centering
        \setlength\tabcolsep{1.25pt}
        \renewcommand{\arraystretch}{1.10}
        \caption{DQN Hyperparameters}
        \vspace{-8pt}
        \begin{tabular}[t]{lc}
            \hline
            \textbf{Hyperparameter} & \textbf{Value}\\
            \hline
            Training steps & 500000\\
            Buffer size & 5000\\
            Optimizer & Adam\\
            Learning rate & $5\times 10^{-3}$\\
            Batch size & 64\\
            N-step Q & 1\\
            Prioritized replay & True\\
            Replay $\alpha$ & 0.6\\
            Replay $\beta$ & 0.4\\
            \hline
        \end{tabular}%
        \label{tab:dqn_hyper_parameters}%
    \end{minipage} 
\end{table}

\subsection{Separate and Ensemble Training (Autoencoder \& Recurrent Parts)}

\begin{figure}[ht!]
	\centering
	\includegraphics[width=\columnwidth]{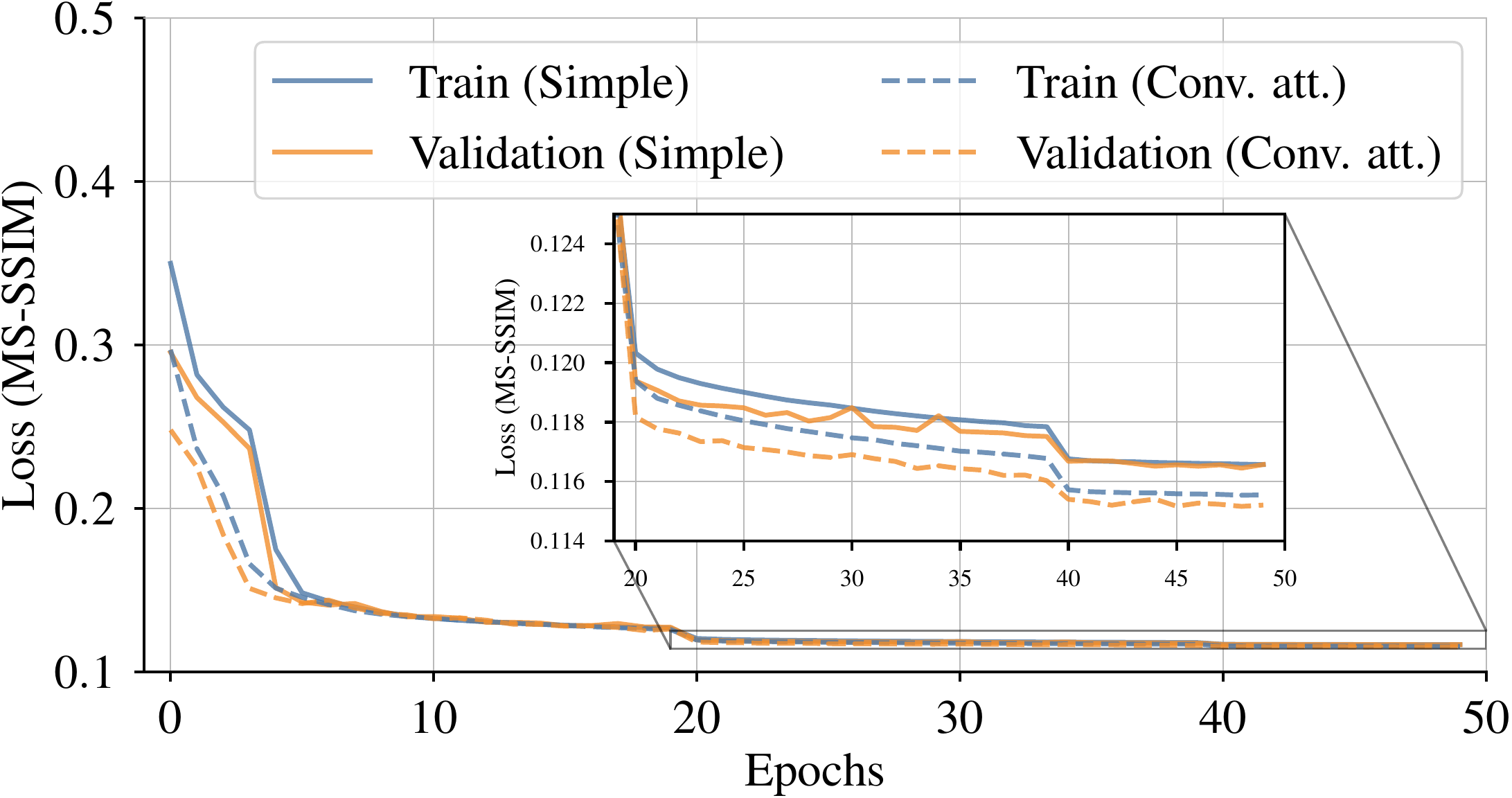}
	\caption{Two autoencoder architectures training results. Adding convolutional self-attention after the image input does not result in any significant qualitative or quantitative improvement in image reconstruction.}
	\label{fig:results_ae_losses}
\end{figure}

\begin{figure}[ht!]
	\centering
	\subfigure[Source (CARLA)]{\includegraphics[width=0.45\columnwidth]{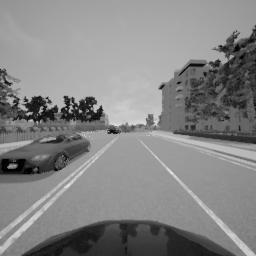}}
	\subfigure[Reconstructed (CARLA)]{\includegraphics[width=0.45\columnwidth]{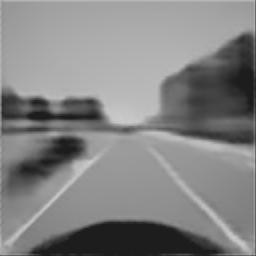}}
	
	\subfigure[Source (Udacity)]{\includegraphics[width=0.45\columnwidth]{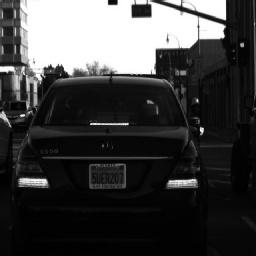}}
	\subfigure[Reconstructed (Udacity)]{\includegraphics[width=0.45\columnwidth]{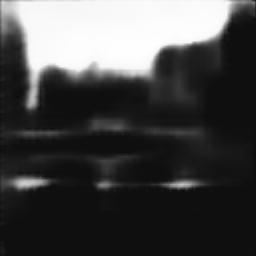}}
	\caption{Autoencoder reconstruction examples. We omit reconstructions for convolution attention-based autoencoder since the difference is imperceptible to human eye.}
	\label{fig:results_ae_reconstruction}
\end{figure}

Recall that the \gls{CARNet} model consists of an autoencoder with a \glsreset{rnn} \gls{rnn} trained together. 
In order to speed up the training process, we pre-trained the autoencoder and fine-tuned the weights as part of the whole network i.e., \gls{ae} + \gls{rnn} (see Fig.~\ref{fig:combined_architecture_all}). 
First, we present the results of the pre-training procedure. 
Note that we also used the convolution attention mechanism (see Fig.~\ref{fig:attention_module}) with the autoencoder model as explained in Section~\ref{sec:cdae_architecture}.
Fig. \ref{fig:results_ae_losses} (latent size 128) shows the training and validation curve for pretrained models, which suggests that there is no over-fitting at the pre-training stage. 
%
%
Example reconstructions for both simulated and real data after pre-training stage are shown in Fig.~\ref{fig:results_ae_reconstruction}.  

\begin{figure*}[!ht]
    \centering
    \subfigure[Source ]{\includegraphics[width=0.19\textwidth]{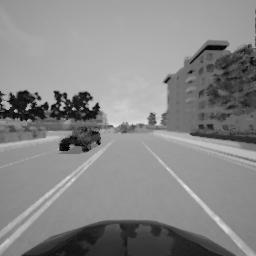}}
    \subfigure[Prediction (LSTM, 128)]{\includegraphics[width=0.19\textwidth]{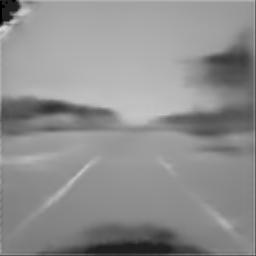}}
    \subfigure[Prediction (GRU, 128)]{\includegraphics[width=0.19\textwidth]{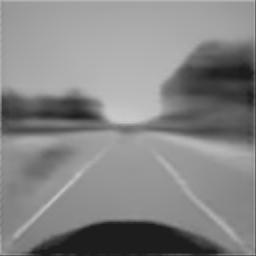}}
    \subfigure[Prediction (LSTM, 64)]{\includegraphics[width=0.19\textwidth]{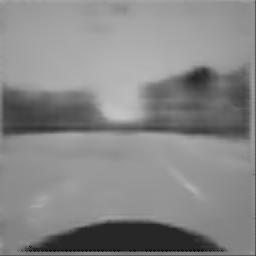}}
    \subfigure[Predcition (GRU, 64)]{\includegraphics[width=0.19\textwidth]{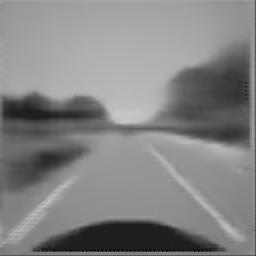}}
    \caption{Comparison of autoencoder reconstruction quality of 512, 128 and 64 latent size vectors. The corresponding training curves are shown in Fig. \ref{fig:results_gru_zoom}.}
    \label{fig:results_ablation_ae_comparison_2}
\end{figure*}

As can be seen in Fig.~\ref{fig:results_ae_reconstruction}, the autoencoder is capable of adequately capturing all necessary parts of the source image. 
Nonetheless, it is important to note that our main aim was to learn latent dynamics rather than ideal image reconstruction.
%
\begin{figure}[ht!]
	\centering
	\includegraphics[width=\columnwidth]{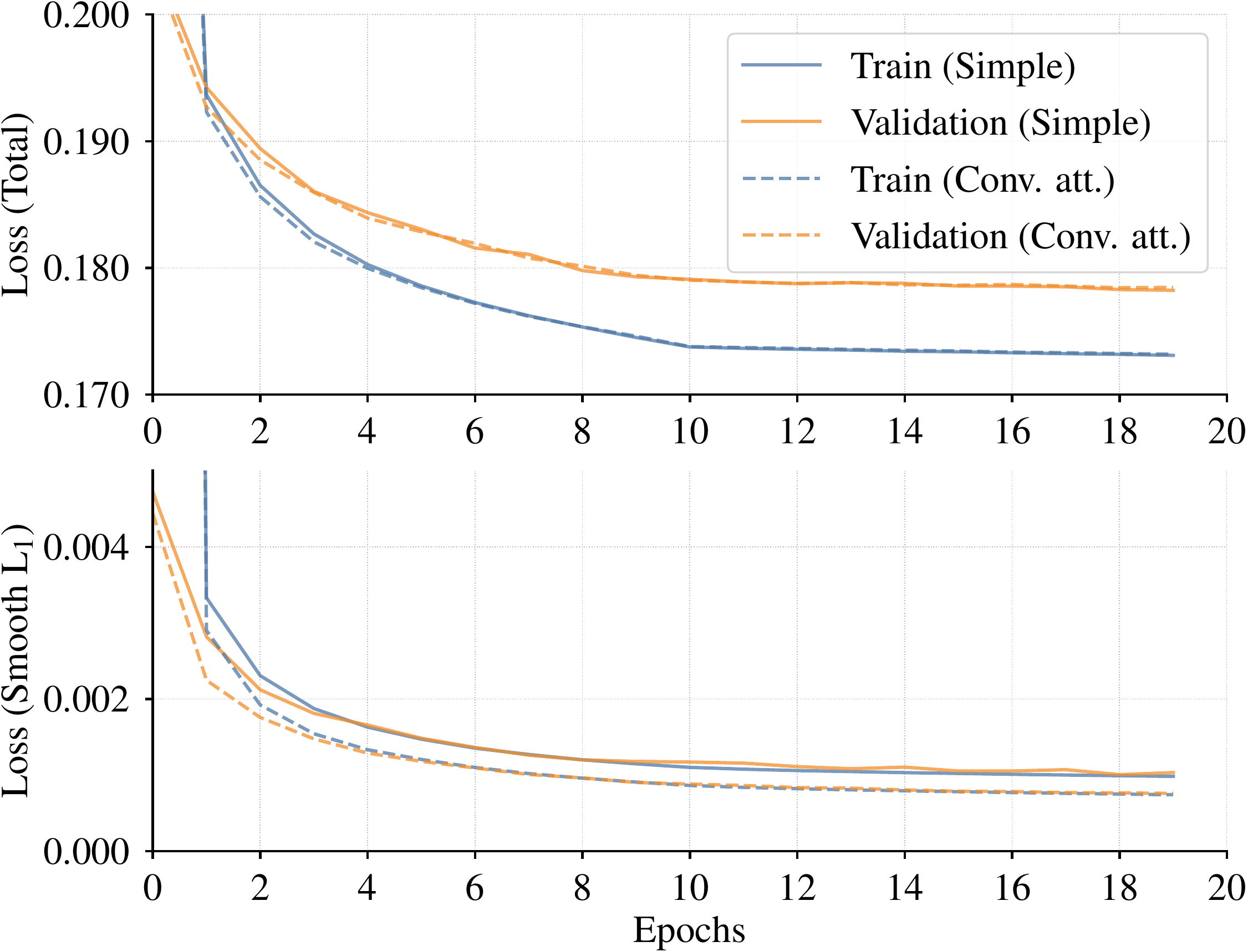} 
	\caption{\gls{gru} architecture training results. Similar to autoencoder part, adding convolutional self-attention did not yield performance improvements. }
	\label{fig:results_gru_zoom}
\end{figure} 
After pre-training the autoencoder, we combined it with the \gls{rnn} and trained them 
together without freezing the weights of the autoencoder.
This is unlike the WM, where each part of the network is trained in isolation in a sequential manner, 
i.e., first the autoencoder, then the recurrent part with autoencoder weights frozen. 
The training losses for the \gls{CARNet} with \gls{gru} are shown in Fig. \ref{fig:results_gru_zoom}. 
The \gls{gru} performed better than \gls{lstm} both in terms of final loss value and overfit. The slight overfit in the image reconstruction appears due to the dataset size. 
Qualitatively, Fig. \ref{fig:results_ablation_ae_comparison_2} shows image reconstructions from predicted latent space.
The quantitative differences in the learned latent variables become more apparent when the imitation learning network is added on top, as shown in the next subsection.

\begin{figure}[ht!]
	\centering
	\includegraphics[width=\columnwidth]{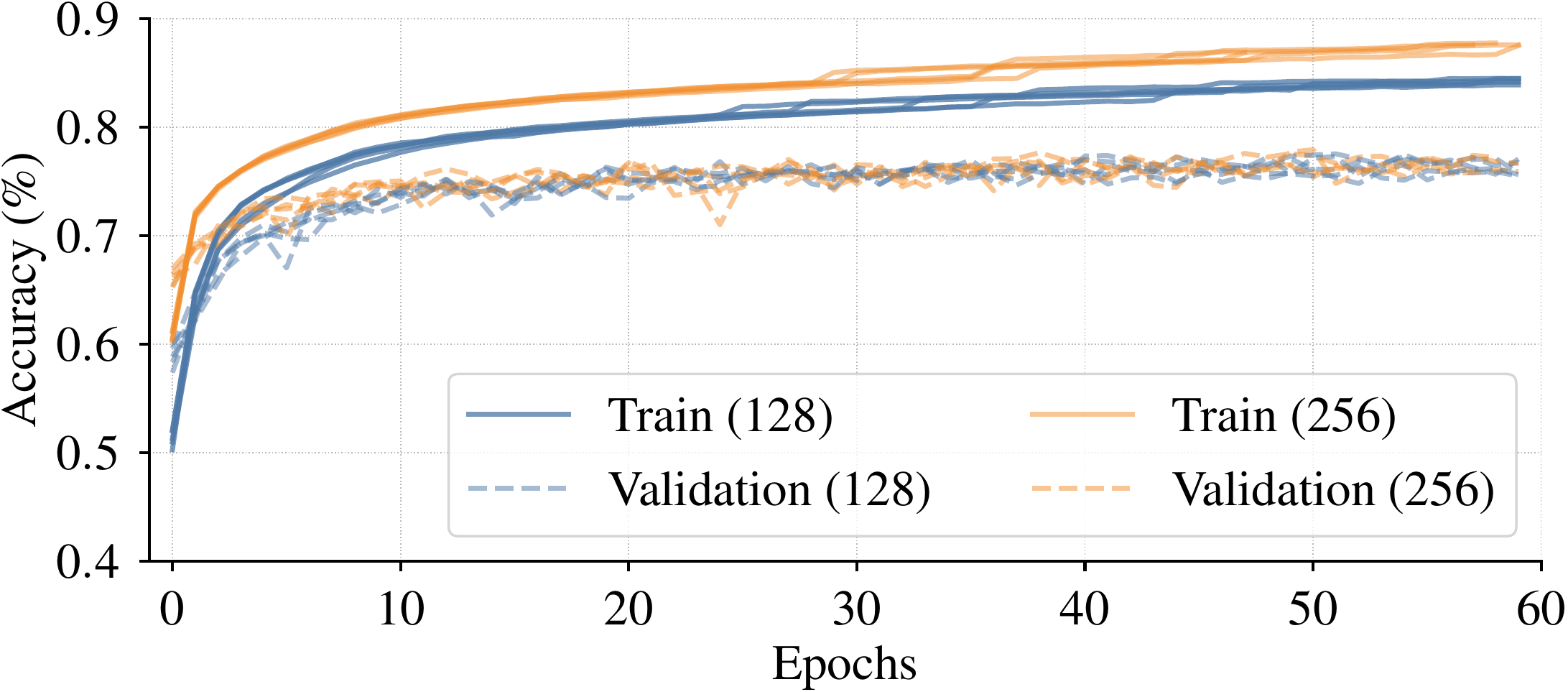}
	\caption{Imitation learning autopilot action classification accuracy. Results are ran five times with resulting accuracies presented Table \ref{tab:results_imitation_latent}.}
	\label{fig:ablation_imitation}
\end{figure}

\subsection{Imitation Learning (CARLA)}

The imitation classification results of the proposed architecture \gls{CARNet} (Fig.~\ref{fig:combined_architecture_all}) and the baseline model (\gls{wm}) is shown in Table~\ref{tab:results_imitation_latent}. 
The \gls{CARNet} had a classification validation accuracy of $\approx$77\%, while \gls{wm} performance was $\approx$71\%.
The proposed \gls{CARNet} model performed significantly better than the World Models (\gls{wm}) while having $\approx$2.5 times fewer parameters. 
Moreover, neither increasing the latent size to 256 (Fig.~\ref{fig:ablation_imitation}) nor adding convolutional self-attention resulted in a noticeable improvement in classification accuracy (Table~\ref{tab:results_imitation_latent}).

\begin{table}[htbp]
\renewcommand{\arraystretch}{1.5}
  \centering
  \caption{Imitation learning results comparison (CARLA simulated dataset).}
    \begin{tabular}{p{40mm}c}
    \hline
    \textbf{Model}  & \textbf{Classification Accuracy\%} \\
    \hline
    \textbf{World Models} & 71.37 $\pm$ 0.52\\
    \textbf{\gls{CARNet} (latent size = 128)} & \textbf{77.08} $\pm$ \textbf{0.51}\\
    \textbf{\gls{CARNet} (latent size = 256)} & \textbf{77.30} $\pm$ \textbf{0.41}\\
    \textbf{\gls{CARNet} (latent size = 128) + Conv. self-attention} & 76.61 $\pm$ 0.39 \\
    \hline
    \end{tabular}%
  \label{tab:results_imitation_latent}%
\end{table}%




\subsection{Imitation Learning (Udacity Dataset)}

\begin{table}[htbp]
\centering
\begin{threeparttable}
\renewcommand{\arraystretch}{1.5}
  \caption{Imitation learning results comparison (Udacity Dataset).}
    \begin{tabular}{p{1.3in}c}
    \hline
    \textbf{Model}  & \textbf{Classification Accuracy\%} \\
    \hline
    \textbf{World Models} & 45.92 $\pm$ 1.66\\
    \textbf{\gls{CARNet} (latent size = 128)} & \textbf{54.24} $\pm$ \textbf{1.24}\\
    \textbf{\gls{CARNet} (latent size = 128), simulation to real\tnote{*}} & \textbf{57.02} $\pm$ \textbf{1.53}\\
    \textbf{\gls{CARNet} (latent size = 256)} & \textbf{59.80} $\pm$ \textbf{1.10}\\
    \hline
    \end{tabular}%
  \label{tab:results_imitation_latent_real}%
  
  \begin{tablenotes}\footnotesize
    \item[*] \added[id=AP]{Network was pre-trained on a simulated and fine-tuned on the real-world dataset.}
    \end{tablenotes}
\end{threeparttable}
\end{table}%

The proposed \gls{CARNet} showed an even better performance margin on real-world data (Udacity dataset). 
While the overall performance accuracies were less than the ones for the simulated dataset, the \gls{CARNet} outperformed \gls{wm} by about 8\% for a latent size of $128$. 
\added[id=AP]{Moreover, in sim-to-real training, when the network was pre-trained on a CARLA dataset and then fine-tuned on the Udacity dataset, the network consistently showed noticeably better performance compared to when trained entirely on real-world data.}
A performance drop on the real dataset is expected since real data is more diverse, e.g., texture, shadows, camera over/under-exposure, etc. 
Additionally, since the minimum possible configuration of the network was explored, the autoencoder might lack the capacity to reconstruct real images compared to simulation quality data. 
While for simulated data there were no performance gains observed when going from a 128 to a 256 latent variable size, for real-world data increasing the latent variable size resulted in about 5\% performance gain as shown in Table \ref{tab:results_imitation_latent_real}.

\subsection{Reinforcement Learning}

For the reinforcement learning task, we followed the same architecture shown in Fig.~\ref{fig:experimental-setup}. 
\begin{table}[htbp]
\renewcommand{\arraystretch}{1.5}
  \centering
  \caption{Reinforcement learning results comparison.}
    \begin{tabular}{lcc}
    \hline
    \textbf{Model}  & \textbf{World Models} & \textbf{\gls{CARNet}} \\
    \hline
    \textbf{Reward} & 2508.77 $\pm$ 249.2 & \textbf{3057.88} $\pm$ \textbf{155.56}\\
    \hline
    \end{tabular}%
  \label{tab:results_reward}%
\end{table}%
The backbone networks (\gls{CARNet} or \gls{wm}) were frozen, and only the fully connected layers were trained. 
However, the fully connected layers were trained using the temporal difference error from the DQN algorithm.
As shown in Table~\ref{tab:results_reward},
the proposed \gls{CARNet} model performed better than \gls{wm}.
The main difference between the proposed \gls{CARNet} model and the \gls{wm} is that the \gls{CARNet} is conditioned on the history of sensor values (i.e, $P(\mathbf{z}_{t+1}|\mathbf{s}_t, \mathbf{z}_t, \mathbf{h}_t)$) rather than the history of actions (i.e, $P(\mathbf{z}_{t+1}|\mathbf{a}_t, \mathbf{z}_t, \mathbf{h}_t)$, see Section~\ref{subs:training}).
Conditioning on the sensor data has two advantages: First, by omitting sensor-action transformations, we avoid the influence of action discretization on the state transition function. 
In other words, if we use the action data, then pre-processing steps such as discretization (from continuous actions) should be applied. 
Depending on the granularity of the action discretization, the quality of the state transition estimation can deteriorate. 
Second, using raw continuous sensor measurements is a more natural way of handling data for autonomous driving, which also avoids additional task-specific post-processing.

\section{Conclusion}

This paper presents a novel approach for learning latent dynamics in an autonomous driving scenario. 
The proposed architecture follows a dynamic convolutional autoencoder structure where the neural network consists of convolutional and recurrent architectures that are trained together. 
The combined architecture provides several advantages, such as reduced trainable parameters and additional continuity constraints on the latent space representation. 
%
%
We showed the efficacy of the proposed model with respect to the established World Models (\gls{wm}) architecure  in both imitation and reinforcement learning scenarios.
Evaluation was performed using both simulated data and real-world driving data.
%
While simpler, the proposed model outperforms the state-of-the-art 
(\gls{wm}) in both imitation learning and reinforcement learning scenarios. 
In imitation learning (cast as a classification problem), the proposed model performance is $\approx 6\%$ and $\approx 8\%$ better than \gls{wm} for simulated real data, respectively. 
In terms of reinforcement learning, the average reward learned through the proposed model is $\approx 549$ (or 22\%) higher than \gls{wm}. 
%
%
%
%
For future work, we aim to explore the generalization and robustness of the proposed model in different scenarios, for example, different town or weather and lighting conditions.

\section*{Acknowledgment}

The authors would like to thank Vlachas Pantelis Rafail for providing valuable advice on recurrent network training and Brian Lee for evaluating the World Models performance on the simulated dataset. We also like to thank Mahdi Ghanei for helping with the synthetic data generation using the CARLA environment.

\bibliographystyle{IEEEtran}
\bibliography{references}


\end{document}